\documentclass[11pt,a4paper]{article}
\usepackage[utf8]{inputenc}
\usepackage[spanish]{babel}
\usepackage{amsmath,amsfonts,amssymb}
\usepackage{graphicx}
\usepackage{hyperref}
\usepackage{enumitem}
\usepackage{geometry}
\usepackage{comment}
\usepackage{tikz}
\usepackage{lscape}
\usepackage{tabularx}

\usepackage{fancyhdr} 
\usetikzlibrary{arrows.meta, positioning, shapes.geometric, shapes, fit, calc}
\addto\captionsspanish{%
}

\tikzset{
  flowStep/.style={rectangle, rounded corners, minimum width=3.2cm, minimum height=1cm, text centered, draw=black, fill=blue!15},
  arrow/.style={thick, ->, >=stealth}
}

\title{\textbf{Procedimiento de auditoría de ciberseguridad para sistemas autónomos: metodología, amenazas y mitigaciones}}

\author{
    Adrián Campazas-Vega, Claudia Álvarez-Aparicio, David Sobrín-Hidalgo, \\
    Laura Inyesto-Alonso, Francisco Javier Rodríguez-Lera, Vicente Matellán-Olivera\\
    y Ángel Manuel Guerrero-Higueras \\
    \small{Grupo de Robótica, Universidad de León} \\
    \small{\texttt{\{acamv, calvaa, dsobh, linya, fjrodl, vicente.matellan, am.guerrero\}@unileon.es}}
}

\date{}

\fancypagestyle{plain}{
  \fancyhf{}
  \fancyhead[C]{\includegraphics[width=\textwidth]{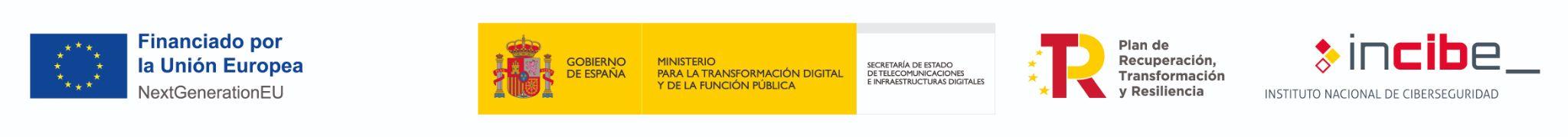}}
  
}
\begin{document}

\maketitle

\begin{abstract}
El despliegue de sistemas autónomos ha experimentado un crecimiento notable en los últimos años, impulsado por su integración en sectores como la industria, la medicina, la logística o el ámbito doméstico. Esta expansión llega acompañada de una serie de problemas de seguridad que adquieren un elevado riesgo debido a la criticidad de los sistemas autónomos, especialmente aquellos que operan en entornos de interacción con humanos. Además, el avance tecnológico y la elevada complejidad operacional y arquitectónica de los sistemas autónomos tiene como consecuencia un aumento en su superficie de ataque. En este artículo se presenta un procedimiento específico de auditoría de seguridad para sistemas autónomos, basado en una metodología estructurada por capas, una taxonomía de amenazas adaptada al contexto robótico y un conjunto de medidas de mitigación concretas. La validez del enfoque propuesto se demuestra mediante cuatro casos prácticos aplicados a plataformas robóticas representativas: el cuadrúpedo militar Vision 60 de Ghost Robotics, el robot A1 de Unitree Robotics, el brazo colaborativo UR3 de Universal Robots y el robot social Pepper de Aldebaran Robotics.
\end{abstract}

\section{Introducción}

Los sistemas autónomos —como robots industriales, drones o vehículos sin conductor— han experimentado un crecimiento sostenido en los últimos años. Según la Federación Internacional de Robótica (IFR), en 2022 se alcanzaron cifras récord en la instalación de robots industriales a nivel global, con un incremento del 290.3 en China, del 39.6\,\% en Estados Unidos y del 3.8\,\% en España~\cite{IFR2023}. Esta expansión no se limita al ámbito productivo: también se extiende a sectores críticos como el transporte, la logística, la medicina y la defensa.

Este avance ha expuesto nuevas vectores de ataque para agentes maliciosos \cite{verma2025systematic}. A diferencia de los sistemas informáticos tradicionales, los sistemas autónomos combinan componentes de hardware, software, sensores, sistemas embebidos y comunicaciones en tiempo real. Esta complejidad los convierte en objetivos de especial interés para ataques dirigidos a interrumpir su funcionamiento, manipular decisiones autónomas o comprometer datos sensibles.

Además de los riesgos técnicos, la ciberseguridad en estos sistemas plantea implicaciones legales y éticas. Es importante considerar que un fallo de seguridad no solo compromete datos o servicios, sino que puede afectar directamente a la seguridad física de personas o instalaciones. Una vulnerabilidad en un robot quirúrgico puede poner en peligro la vida de un paciente; un dron comprometido podría utilizarse con fines hostiles. En consecuencia, resulta imprescindible desarrollar procedimientos de análisis de vulnerabilidades o auditorías de seguridad  adaptados a las características particulares de los sistemas autónomos.

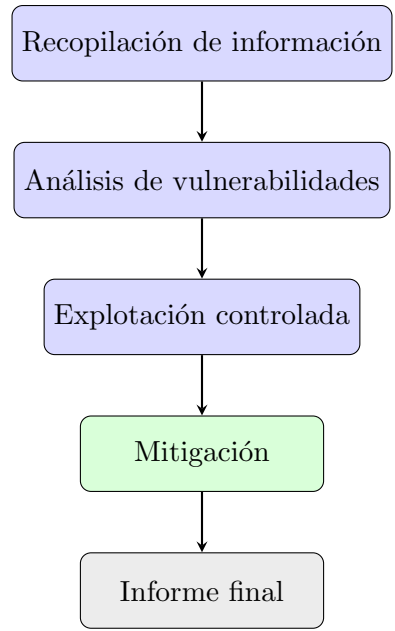
\begin{figure}[h]
\centering
\begin{tikzpicture}[node distance=0.8cm]
\node (info) [flowStep] {Recopilación de información};
\node (analisis) [flowStep, below=of info] {Análisis de vulnerabilidades};
\node (exploit) [flowStep, below=of analisis] {Explotación controlada};
\node (mitigacion) [flowStep, below=of exploit, fill=green!15] {Mitigación};
\node (reporte) [flowStep, below=of mitigacion, fill=gray!15] {Informe final};

\draw [arrow] (info) -- (analisis);
\draw [arrow] (analisis) -- (exploit);
\draw [arrow] (exploit) -- (mitigacion);
\draw [arrow] (mitigacion) -- (reporte);
\end{tikzpicture}
\caption{Flujo del procedimiento de auditor\'ia de seguridad para sistemas aut\'onomos.}
\label{fig:flujo_auditoria}
\end{figure}

Este trabajo presenta una propuesta metodológica de auditoría de seguridad orientada específicamente a este tipo de plataformas. El procedimiento se articula en tres fases --recopilación de información, análisis de vulnerabilidades y explotación controlada-- y se apoya en un modelo de análisis por capas, una taxonomía de amenazas organizada por dominios tecnológicos, y una selección de medidas de mitigación concretas. La Figura ~\ref{fig:flujo_auditoria} presenta el flujo de trabajo del procedimiento de auditoría propuesto. La aplicabilidad del enfoque se valida mediante casos de uso prácticos sobre las plataformas Vision 60 de Ghost Robotics, el robot cuadrúpedo Unitree A1, el brazo robótico UR3 de Universal Robots y el robot social Pepper de Aldebaran Robotics, ilustrando las capacidades del procedimiento para detectar y evaluar riesgos reales.

\section{Metodología de auditoría de seguridad para sistemas autónomos}

Auditar la seguridad de sistemas autónomos plantea retos específicos que van más allá de los enfoques tradicionales aplicados en entornos informáticos. La diversidad de componentes —sensores, actuadores, sistemas operativos embebidos, buses de comunicación, etc.— y la criticidad de sus aplicaciones hacen necesaria una metodología adaptada a este tipo de sistemas.

De esta forma, la metodología propuesta para hacer frente a la complejidad subyacente a los sistemas autónomos se estructura en tres fases:
\begin{enumerate}
    \item \textbf{Recopilación de información}.
    \item \textbf{Análisis de vulnerabilidades}.
    \item \textbf{Explotación controlada}.
\end{enumerate}

Esta estrategia se inspira en marcos de referencia como IEC 62443~\cite{IEC62443}, aplicable a sistemas industriales, y el enfoque de auditoría para robots propuesto por Yaacoub et al.~\cite{yaacoub2022robotics} y por Mayoral et al.~\cite{2018arXiv180604042M}, pero se adapta a las características dinámicas y heterogéneas de los sistemas autónomos.

\subsection{Recopilación de información}

La fase inicial de la auditoría se centra en obtener una visión estructurada del sistema objetivo. El objetivo principal es delimitar su \emph{superficie de ataque}, es decir, el conjunto de posibles puntos de entrada susceptibles de ser utilizados por un adversario para comprometer el sistema.

Esta etapa incluye cinco tareas clave:

\begin{itemize}
  \item \textbf{Inventario de hardware}: identificación de sensores, actuadores, controladores, módulos de comunicación (Wi-Fi, 5G, Bluetooth, LoRa, etc.) y medios de almacenamiento. Esta información permite anticipar posibles vectores de ataque físico.

  \item \textbf{Topología de red}: elaboración de un mapa detallado de las conexiones internas y externas del sistema. Incluye interfaces inalámbricas, buses como I2C o CAN, y enlaces con estaciones de control o servicios en la nube.

  \item \textbf{Software y firmware}: recopilación de versiones del sistema operativo (e.g., Ubuntu Core, RTOS), frameworks robóticos (ROS, ROS 2), bibliotecas, firmware de bajo nivel y módulos personalizados.

  \item \textbf{Servicios expuestos}: identificación de servicios activos, como puertos abiertos, APIs REST/GraphQL, interfaces SSH o Telnet, y servicios web accesibles desde la red.

  \item \textbf{Mecanismos de autenticación y cifrado}: revisión del uso de protocolos seguros (TLS, SSH), autenticación multifactor, certificados digitales, y presencia de módulos de seguridad hardware (HSM o TEE).

  \item \textbf{Aplicaciones externas}: revisión de las aplicaciones de control del dispositivo como por ejemplo una APK desarrollada para el sistema operativo Android, la información relevante que contiene  y las comunicaciones que efectua con el robot. 
\end{itemize}

Es importante destacar el aumento significativo que la superficie de ataque puede sufrir si un robot es manejado por una aplicación externa. El uso de una aplicación de este tipo supone la realización completa de una auditoría de seguridad en dicha aplicación ya que un problema de seguridad en dicha utilidad puede suponer un problema de seguridad en el robot. A modo de ejemplo la Figura \ref{fig:auditoria_android}, detalla la superficie de ataque de una aplicación móvil basada en la metodología OWASP \cite{owasp2025} y en su Top 10 de vulnerabilidades que se detalla a continuación:

\begin{figure}[h!]
\centering
\includegraphics[width=0.85\textwidth]{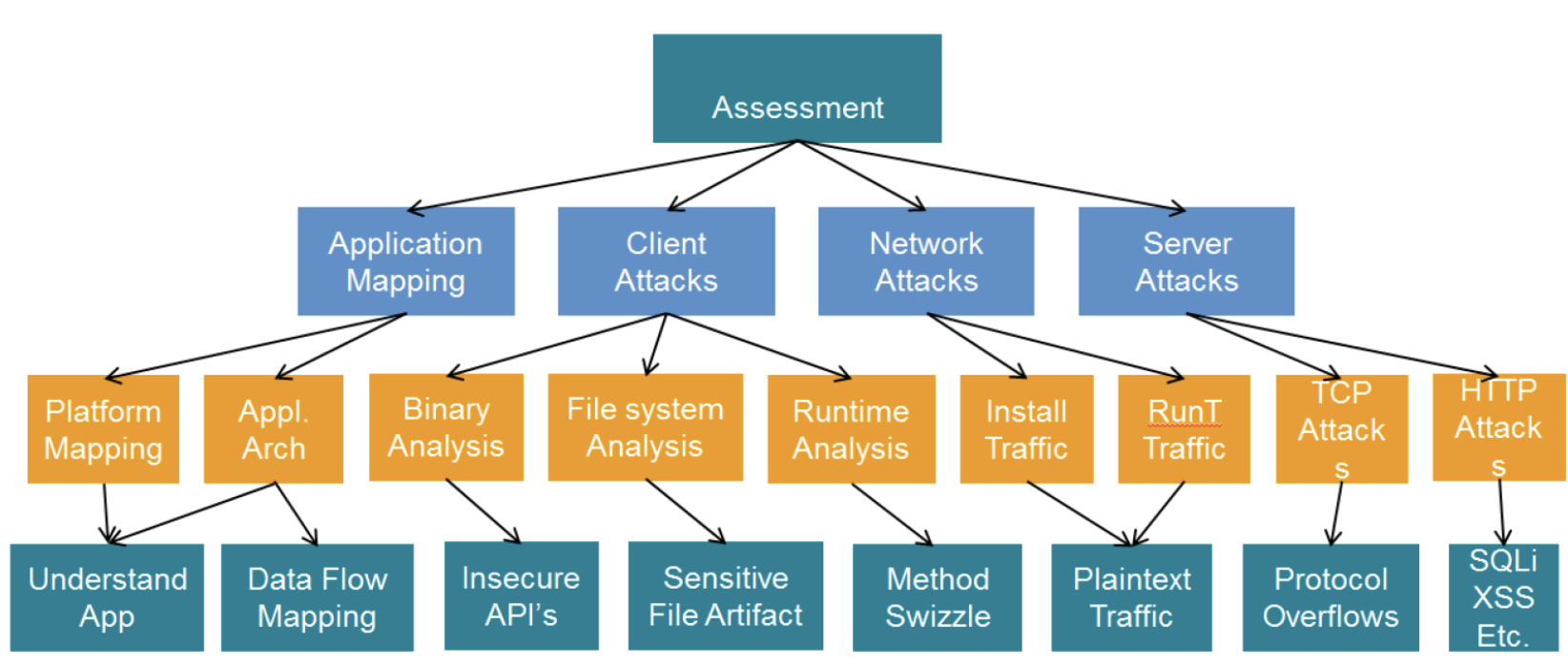}
\caption{Metodología Auditoría Apps móviles}
  \label{fig:auditoria_android}
\end{figure}

\begin{itemize}[leftmargin=1.5em]

\item \textbf{M1 - Uso inadecuado de la plataforma:} Ocurre cuando una app no utiliza correctamente las APIs del sistema operativo móvil, lo que puede llevar a vulnerabilidades como mal uso del almacenamiento, permisos inseguros o configuraciones incorrectas.

\item \textbf{M2 - Almacenamiento inseguro de datos:} Se refiere a cuando los datos sensibles del usuario (como contraseñas o tokens) se guardan sin la debida protección, permitiendo que sean accedidos por apps maliciosas o usuarios con acceso físico.

\item \textbf{M3 - Comunicación insegura:} Involucra la transmisión de datos sin cifrado o con cifrado débil, lo que permite ataques como el "man-in-the-middle", especialmente en redes públicas.

\item \textbf{M4 - Autenticación insegura:} Problemas en cómo la app verifica la identidad del usuario, como uso de contraseñas débiles, falta de autenticación de dos factores o gestión inadecuada de sesiones.

\item \textbf{M5 - Insuficiente criptografía:} Se refiere al uso incorrecto o débil de algoritmos de cifrado, como claves demasiado cortas, algoritmos obsoletos o mal uso de librerías criptográficas.

\item \textbf{M6 - Autorización insegura:} Se da cuando una app no valida correctamente los privilegios del usuario, permitiéndole realizar acciones o acceder a recursos que no debería.

\item \textbf{M7 - Calidad del código cliente:} Implica prácticas de desarrollo pobres como código ofuscado débil, errores lógicos, o dejar funciones de depuración activas, lo cual facilita la ingeniería inversa o explotación de fallos.

\item \textbf{M8 - Ingeniería inversa:} Ocurre cuando los atacantes analizan la app mediante decompiladores o herramientas similares para extraer información sensible o encontrar vulnerabilidades.

\item \textbf{M9 - Manipulación de código:} Se refiere a modificar el código de la app (por ejemplo, con apps ''crackeadas'') para alterar su comportamiento, eliminar restricciones o inyectar funciones maliciosas.

\item \textbf{M10 - Gestión de la cadena de suministro:} Se centra en la seguridad de librerías externas y componentes de terceros que la app utiliza, ya que una dependencia comprometida puede afectar a toda la aplicación.

\end{itemize}

Para realizar las tareas enumeradas en esta sección, se recomienda el uso de herramientas especializadas como \texttt{Nmap}, \texttt{Wireshark} o técnicas de \emph{fuzzing} orientadas a interfaces robóticas. La documentación generada permite construir un modelo del sistema desde la perspectiva del atacante.

La Tabla~\ref{tab:tools_recon} resume las herramientas más útiles para cada una de estas tareas, junto con ejemplos de uso.

\begin{table}[h]
\centering
\caption{Herramientas recomendadas para la recopilación de información}
\label{tab:tools_recon}
\begin{tabular}{|p{5cm}|p{7.5cm}|}
\hline
\textbf{Tarea} & \textbf{Herramientas sugeridas} \\
\hline
Inventario de hardware &
\texttt{lshw}, \texttt{lsusb}, \texttt{lspci}, \texttt{i2cdetect}, \texttt{dmidecode} \\
\hline
Topología de red &
\texttt{ip a}, \texttt{ip route}, \texttt{nmap}, \texttt{arp-scan}, \texttt{netstat} \\
\hline
Software y firmware &
\texttt{uname -a}, \texttt{lsb\_release -a}, \texttt{binwalk} \\
\hline
Servicios expuestos &
\texttt{nmap}, \texttt{ss}, \texttt{netstat}, \texttt{shodan}, \texttt{masscan} \\
\hline
Autenticación y cifrado &
\texttt{sslscan}, \texttt{testssl.sh}, \texttt{Wireshark} \\
\hline
Aplicaciones externas &
\texttt{apktool}, \texttt{frida}, \texttt{dex2jar}\\
\hline
\end{tabular}
\end{table}

\subsection{Análisis de vulnerabilidades}

Una vez documentada la arquitectura del sistema, se procede a identificar debilidades explotables en sus distintas capas. Esta fase combina técnicas manuales y herramientas automatizadas, y requiere un enfoque multidisciplinar debido a la naturaleza ciberfísica de los sistemas autónomos, donde las vulnerabilidades pueden afectar tanto a componentes digitales como físicos.

Las principales actividades se agrupan en cuatro dominios:

\begin{itemize}
  \item \textbf{Hardware}~\cite{bhunia2014hardware, tehranipoor2011hardware}: se analiza la presencia de interfaces de depuración activas (UART, JTAG), componentes expuestos físicamente o susceptibles de manipulación electromagnética, así como la posible inserción de troyanos en la cadena de suministro. También se analizan los accesos a puertos de control como USB sin mecanismos de autenticación.

  \item \textbf{Firmware y sistema operativo}: mediante técnicas de ingeniería inversa se extraen y analizan imágenes de firmware. Se buscan configuraciones inseguras, contraseñas embebidas, servicios innecesarios y ausencia de mecanismos como arranque seguro o aislamiento de privilegios. También se analizan las vulnerabilidades intrínsecas del propio sistema operativo sobre el que se despliega el robot.

  \item \textbf{Software de control}: se auditan las aplicaciones de usuario (como apps móviles o interfaces web), prestando especial atención al uso de bibliotecas obsoletas, almacenamiento inseguro de credenciales y validaciones insuficientes en la comunicación con el sistema.

  \item \textbf{Redes y comunicaciones}: se analiza el tráfico generado para detectar protocolos no cifrados, ausencia de autenticación, y susceptibilidad a ataques como replay, spoofing o \emph{Man-in-the-Middle} (MitM). También se estudian protocolos propietarios potencialmente inseguros.
\end{itemize}

Es importante aclarar que estas actividades deben desarrollarse en entornos controlados para evitar daños al sistema o a su entorno. El análisis debe estar documentado y respetar los principios de auditoría ética.

La Tabla~\ref{tab:vuln_tools} recoge las herramientas más relevantes para cada tipo de análisis, junto con técnicas y ejemplos de uso habituales en entornos reales.

\begin{table}[h]
\centering
\caption{Herramientas para el análisis de vulnerabilidades en sistemas autónomos}
\label{tab:vuln_tools}
\begin{tabular}{|p{4.5cm}|p{8cm}|}
\hline
\textbf{Área} & \textbf{Herramientas sugeridas} \\
\hline
Hardware &
\texttt{JTAGulator}, \texttt{OpenOCD}, osciloscopios, interferencia EM \\
\hline
Firmware y S.O. &
\texttt{binwalk}, \texttt{firmware-mod-kit}, \texttt{QEMU}, \texttt{chkrootkit} \\
\hline
Software de control &
\texttt{MobSF}, \texttt{Ghidra}, \texttt{apktool}, \texttt{Bandit} (Python) \\
\hline
Red y comunicaciones &
\texttt{Wireshark}, \texttt{ettercap}, \texttt{mitmproxy}, \texttt{zmap} \\
\hline
\end{tabular}
\end{table}

\subsection{Explotación controlada de vulnerabilidades}

Una vez identificadas y clasificadas las vulnerabilidades potenciales del sistema detectadas en la fase anterior, es necesario verificar su impacto real mediante técnicas de explotación controlada. Esta fase tiene como objetivo confirmar la explotabilidad de las debilidades detectadas y observar sus efectos en el comportamiento del sistema autónomo.

A diferencia de pruebas destructivas, la explotación controlada se realiza en entornos aislados o simulados, con el propósito de estudiar el alcance de cada fallo sin comprometer la integridad física o funcional del sistema en producción.

Las principales actividades de explotación pueden agruparse en correspondencia con los dominios analizados en la fase previa:

\begin{itemize}
\item \textbf{Hardware}: se realizan intentos de acceso a través de interfaces de depuración expuestas (UART, JTAG), conexión de periféricos maliciosos en puertos de control (p.~ej., USB) o inducción de fallos físicos que permitan evadir mecanismos de protección.

\item \textbf{Firmware y sistema operativo}: se llevan a cabo pruebas de escalada de privilegios explotando configuraciones inseguras o vulnerabilidades conocidas del kernel, utilización de credenciales embebidas para obtener control administrativo, y verificación de la posibilidad de cargar firmware modificado con el fin de desactivar medidas de seguridad como el arranque seguro.

\item \textbf{Software de control}: se examina la explotación de aplicaciones de usuario (interfaces web o móviles) mediante inyección de comandos, aprovechamiento de bibliotecas obsoletas o acceso a credenciales almacenadas de manera insegura, lo que podría facilitar la persistencia de un atacante en el sistema.

\item \textbf{Redes y comunicaciones}: se evalúa la posibilidad de sufrir ataques de interceptación (Man-in-the-Middle) y repetición (replay) en canales no cifrados, la inyección de datos falsos en mensajes de control, así como las denegaciones de servicio dirigidas contra protocolos críticos.
\end{itemize}

Este tipo de pruebas es crucial para evaluar las consecuencias en cascada de una vulnerabilidad: cómo una brecha localizada (por ejemplo, una contraseña débil en una API) puede derivar en el control completo del sistema, afectando a módulos críticos como el firmware, los actuadores o los sistemas de comunicación. Este enfoque sigue el principio de \emph{seguridad en profundidad}, ampliamente promovido en el diseño de arquitecturas ciberfísicas robustas \cite{humayed2017cyberphysical}.

En la Tabla~\ref{tab:exploitation_tools} se muestran herramientas y técnicas frecuentemente utilizadas para llevar a cabo este tipo de pruebas, junto con ejemplos representativos de su aplicación.

\begin{table}[h]
\centering
\caption{Herramientas para explotación controlada de vulnerabilidades}
\label{tab:exploitation_tools}
\begin{tabular}{|p{4.5cm}|p{8cm}|}
\hline
\textbf{Objetivo de prueba} & \textbf{Herramientas sugeridas} \\
\hline
Acceso no autorizado &
\texttt{Hydra}, \texttt{Medusa}, \texttt{ADB}, \texttt{telnet}, \texttt{screen} \\
\hline
Intercepción/Modificación de comandos &
\texttt{mitmproxy}, \texttt{Ettercap}, \texttt{Wireshark}, \texttt{Scapy} \\
\hline
Escalada de privilegios &
\texttt{LinPEAS}, \texttt{GTFOBins}, \texttt{sudo -l}, \texttt{ExploitDB} \\
\hline
Inyección de datos falsos (FDI) &
\texttt{Scapy}, \texttt{ROS2 CLI}, scripts personalizados \\
\hline
Denegación de servicio &
\texttt{hping3}, \texttt{Slowloris}, \texttt{LOIC}, scripts de fuzzing \\
\hline
\end{tabular}
\end{table}

\paragraph{Consideraciones éticas y responsabilidad legal}

Toda actividad de explotación debe desarrollarse en entornos controlados, con autorización explícita de los propietarios del sistema y sin afectar servicios productivos. La auditoría ética se rige por principios de proporcionalidad, minimización del daño y respeto a la privacidad. Además, es responsabilidad del equipo auditor cumplir con la legislación vigente en materia de ciberseguridad, como el Reglamento General de Protección de Datos (RGPD)~\cite{blanco2018reglamento} en Europa o normativas sectoriales aplicables (como ISO/IEC 27001~\cite{ISOGuide72} o la Directiva NIS2~\cite{EU-Directive-2022-2555}). Las pruebas deben ser debidamente documentadas y justificadas, especialmente en entornos críticos como sanidad, defensa o transporte autónomo. Además se debe establecer un convenio entre la entidad propietaria del sistema y el auditor en el que se establezcan las fechas y las horas concretas en las cuales se realizaran las pruebas, de forma que la explotación de dichos entornos no tenga un impacto negativo en el cometido diario de la entidad propietaria.

\subsection{Aplicación práctica de la estrategia de análisis por capas}

La complejidad estructural y funcional de los sistemas autónomos modernos, especialmente aquellos desplegados en entornos críticos o dinámicos, exige una metodología de auditoría que permita un análisis exhaustivo y sistemático. En este contexto, la estrategia de \emph{análisis por capas} ha demostrado ser particularmente eficaz.

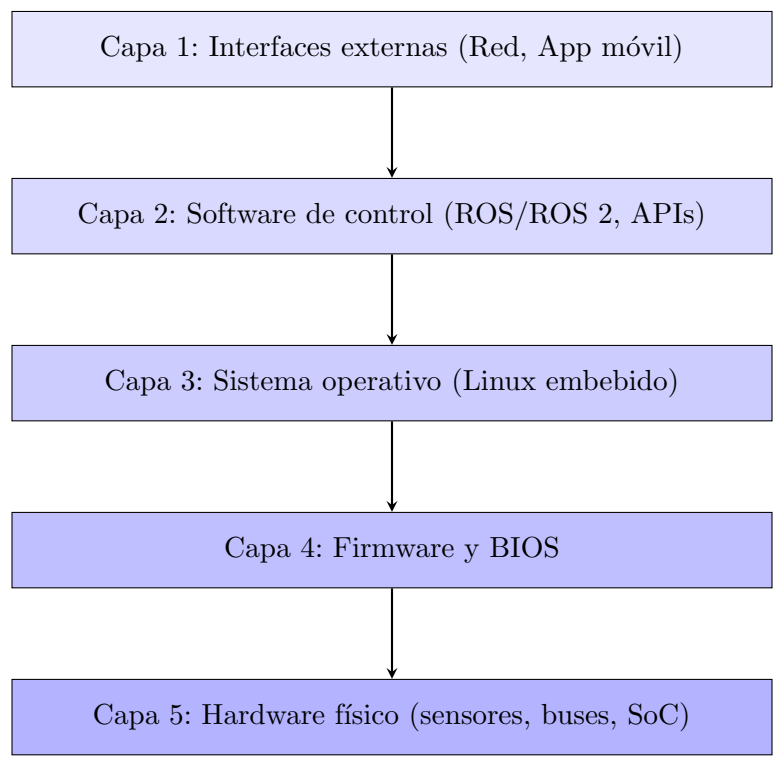
\begin{figure}[h]
\centering
\begin{tikzpicture}[
  node distance=1.2cm,
  every node/.style={minimum width=10cm, minimum height=1cm, text centered},
  box/.style={draw, fill=#1, minimum height=1cm}
]

% Capas
\node[box=blue!10] (c1) {Capa 1: Interfaces externas (Red, App móvil)};
\node[box=blue!15, below=of c1] (c2) {Capa 2: Software de control (ROS/ROS 2, APIs)};
\node[box=blue!20, below=of c2] (c3) {Capa 3: Sistema operativo (Linux embebido)};
\node[box=blue!25, below=of c3] (c4) {Capa 4: Firmware y BIOS};
\node[box=blue!30, below=of c4] (c5) {Capa 5: Hardware físico (sensores, buses, SoC)};

% Flechas
\draw [arrow] (c1) -- (c2);
\draw [arrow] (c2) -- (c3);
\draw [arrow] (c3) -- (c4);
\draw [arrow] (c4) -- (c5);
\end{tikzpicture}
\caption{Modelo de análisis por capas en sistemas autónomos.}
\label{fig:capas_auditoria}
\end{figure}

Este enfoque propone examinar el sistema desde el exterior hacia el interior, comenzando por las interfaces más expuestas (como comunicaciones inalámbricas, APIs, y aplicaciones móviles), y avanzando progresivamente hacia los componentes de control interno, el sistema operativo, el firmware y, finalmente, el hardware y los sensores físicos. Esta progresión descendente permite identificar vectores de ataque en fases tempranas y establecer rutas potenciales de explotación que atraviesen múltiples capas del sistema. La Figura~\ref{fig:capas_auditoria} ilustra este modelo de auditoría por capas aplicado a sistemas autónomos.

Una ventaja clave de esta estrategia es que prioriza la superficie de ataque efectiva, es decir, los puntos más accesibles y probables de ser atacados, sin descuidar elementos internos que pueden verse comprometidos en ataques en cadena. Además, facilita la segmentación del trabajo entre equipos de auditoría especializados (por ejemplo, redes, sistemas embebidos, desarrollo móvil) y promueve la documentación estructurada del análisis.

El enfoque planteado se ha validado en \cite{paperUnitreeA1, campazas2023cybersecurity}, dos trabajos centrados en el proceso de auditoría sobre plataformas industriales y robots autónomos, donde la organización modular de los sistemas hace especialmente eficaz una aproximación escalonada.

\subsection{Análisis de la criticidad de las vulnerabilidades}

Para evaluar la criticidad de las vulnerabilidades descubiertas durante el proceso de auditoría se propone el uso de Common Vulnerability Scoring System (CVSS) version 3 \cite{incibe2023cvss}. CVSS es un framework que define una serie de métricas para identificar las características, el impacto y la gravedad de las vulnerabilidades. A través de CVSSv3 se asigna un valor numérico a la vulnerabilidad de forma que se construyen las siguientes categorías:

\begin{itemize}
    \item \textbf{Gravedad nula}: puntuación de 0
    \item \textbf{Gravedad baja}: puntuación entre 0.1 y 3.9
    \item \textbf{Gravedad media o moderada}: puntuación entre 4.0 y 6.9
    \item \textbf{Gravedad alta}: puntuación entre 7.0 y 8.9
    \item \textbf{Gravedad crítica}: puntuación entre 9.0 y 10
\end{itemize}

El sistema de puntuación descrito permite categorizar la gravedad de las vulnerabilidades de forma que facilite los esfuerzos para prevenir y mitigar la ocurrencia de ataques que explotan dichas vulnerabilidades.

\section{Validación de la metodología: casos reales}

Para evaluar la aplicabilidad práctica del procedimiento de auditoría propuesto en este trabajo, se ha llevado a cabo su implementación sobre cuatro plataformas robóticas comerciales con arquitectura ciberfísica compleja: el robot cuadrúpedo \textit{Vision 60} de Ghost Robotics, el modelo \textit{A1} de Unitree Robotics, el brazo robótico colaborativo \textit{UR3} de Universal Robots y el robot social Pepper de Aldebaran Robotics. Estas plataformas, aunque diversas en diseño y propósito —desde locomoción autónoma en entornos adversos hasta manipulación de precisión en entornos industriales— comparten características clave como conectividad remota, integración de sensores, sistemas embebidos y control mediante interfaces de red. Dichas propiedades las convierten en casos representativos para validar tanto la metodología propuesta.

Cada estudio ha seguido de forma rigurosa el esquema metodológico propuesto, abarcando la recopilación de información técnica, la identificación de vectores de ataque, la evaluación de vulnerabilidades, la explotación controlada en un entorno de laboratorio y la formulación de recomendaciones específicas de mitigación. A continuación, se detallan los resultados obtenidos para cada plataforma, estructurados por fases.

\subsection{Caso 1: Vision 60 de Ghost Robotics}

\begin{figure}[htbp]
    \centering
    \includegraphics[width=0.5\linewidth]{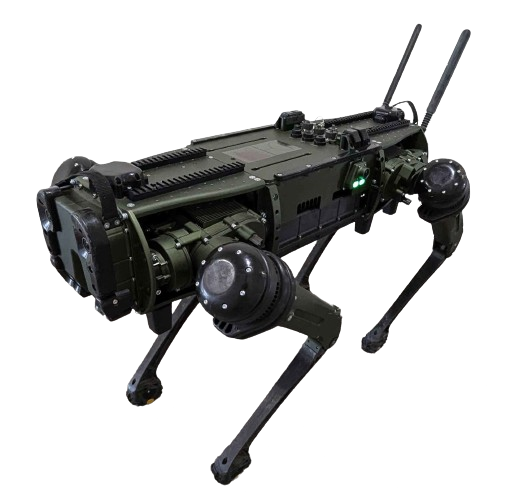}
    \caption{Vision 60 de Ghost Robotics}
    \label{fig:ghost}
\end{figure}

El Vision 60 (ver Figura~\ref{fig:ghost}) es un robot cuadrúpedo militar de alta capacidad desarrollado por Ghost Robotics, orientado a operaciones en exteriores, vigilancia autónoma y despliegues en condiciones adversas. Su arquitectura se basa en una distribución de ROS 2 \cite{ros2} sobre un sistema Ubuntu. ROS 2 se considera actualmente el estandar de facto en lo referente al desarrollo de aplicaciones robóticas. Supone la evolución de su predecesor (ROS) e incorpora importantes cambios desde una perspectiva de ciberseguridad como la integración de mecanismos de autenticación y cifrado.

El uso de esta tecnología también abre vectores de ataque comunes en sistemas ROS mal configurados. El análisis reveló importantes debilidades en las interfaces de control, las políticas de autenticación y el cifrado de datos.

\begin{itemize}
  \item \textbf{Recopilación de información:} Esta fase se ha centrado en obtener el máximo conocimiento sobre la arquitectura del sistema sin realizar modificaciones intrusivas. Se han inspeccionado los componentes físicos, el sistema operativo, los servicios en ejecución y los canales de comunicación.
  \begin{itemize}
    \item Se detectaron interfaces de comunicación físicas y lógicas, incluyendo conectores USB-C, un módulo WiFi con conectividad directa, y conectores Ethernet (ver Figura~\ref{fig:ghost_puertos}).
    \item Se identificó que el sistema operativo se corresponde con un Ubuntu 18.04 que contiene una instalación de ROS 2 Dashing, con servicios como SSH, HTTP y ROS DDS Discovery en puertos dinámicos.
    \item Se detectó que la administración del robot se realiza a nivel de API, utilizando  utilizando comunicaciones  HTTP sin cifrado y sin autenticación por defecto.
  \end{itemize}

\begin{figure}[htbp]
    \centering
    \includegraphics[width=0.5\linewidth]{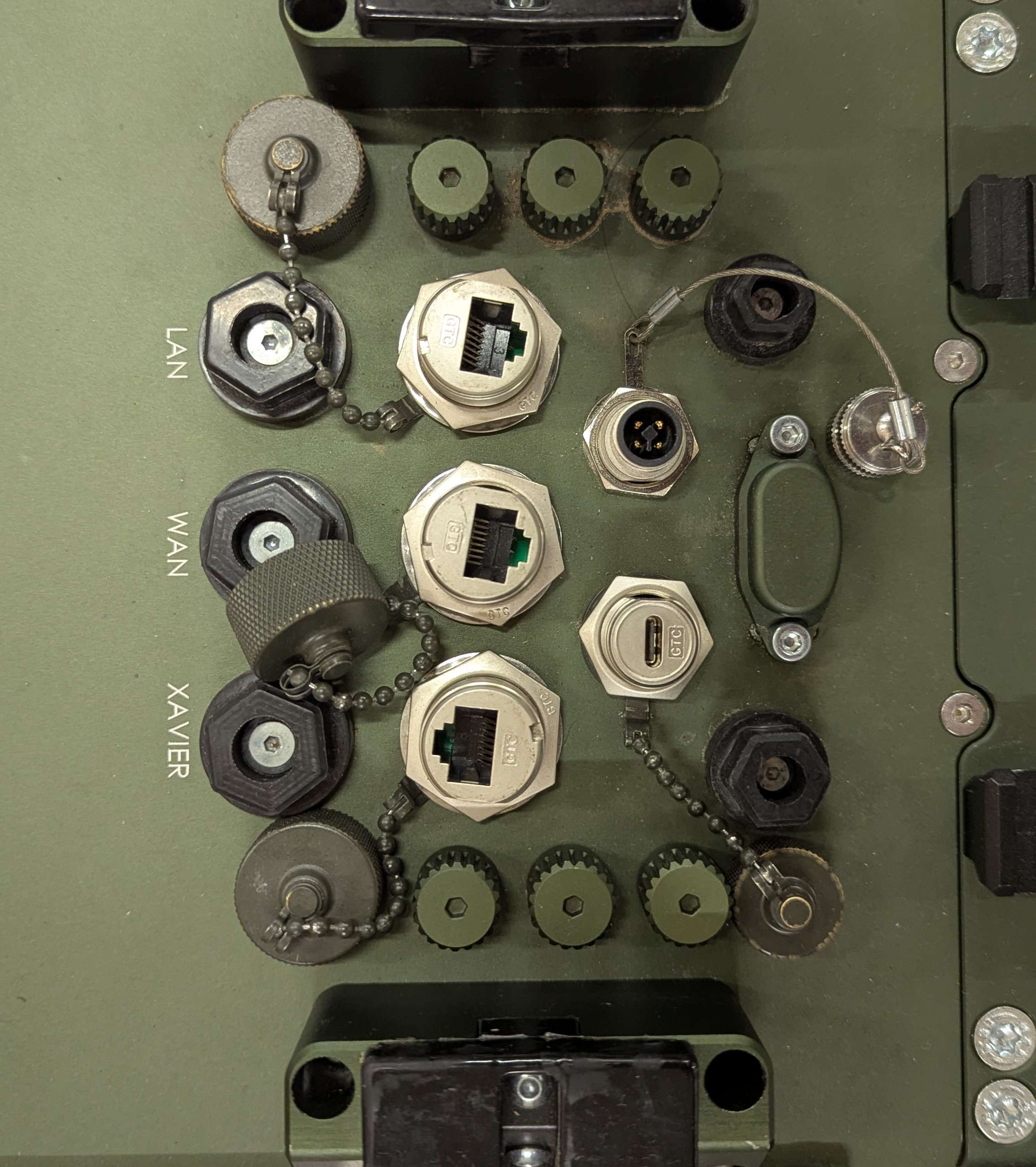}
    \caption{Conexiones superiores del Vision 60}
    \label{fig:ghost_puertos}
\end{figure}

  \item \textbf{Análisis de vulnerabilidades:} Aplicando herramientas como Lynis, OpenVAS y scripts personalizados, se han analizado debilidades correspondientes tanto al sistema operativo como al middleware.
  \begin{itemize}
    \item El uso de ROS 2 sin configuración de seguridad (como SROS2) permite la suscripción e inyección directa de mensajes en los topics del robot.
    \item Se han hallado credenciales codificadas en la aplicación cliente, asociadas a los parámetros de conexión del sistema (ver Figura~\ref{fig:ghost_texto_plano}).
    \item Se detectó que la consola web carece de un sistema de autenticación.
    \item El sistema no implementa políticas de control de acceso ni segmentación de red, lo que permite movimientos laterales entre servicios.
    \item No se implementan políticas de control de acceso a nivel de puerto físico en el dispositivo.
  \end{itemize}

\begin{figure}[htbp]
    \centering
    \includegraphics[width=0.5\linewidth]{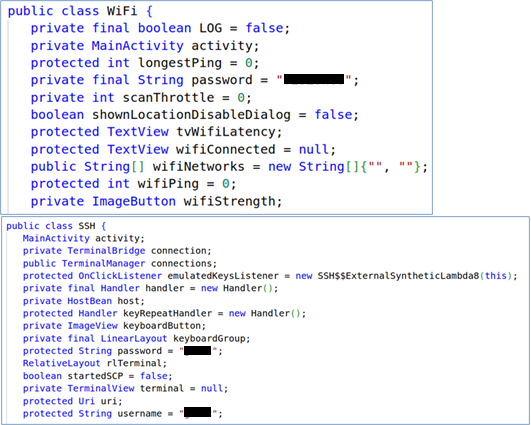}
    \caption{Fragmentos del código decompilado de la APK donde se puede
observar en texto plano las contraseñas y el usuario}
    \label{fig:ghost_texto_plano}
\end{figure}

  \item \textbf{Explotación controlada:} Las vulnerabilidades encontradas se probaron en un entorno controlado, con el fin de observar su impacto real sin comprometer la integridad del sistema.
  \begin{itemize}
    \item Se ha simulado un ataque DoS sobre el middleware ROS 2 enviando mensajes malformados con una elevada frecuencia de publicación a través del protocolo DDS, lo que ha provocado latencias y pérdida de paquetes en temas de control.
    \item Se ha explotado el sistema distribuido de nodos y topics de ROS 2 mediante una inyección directa de parámetros que ha permitido modificar valores internos de la locomoción del robot.
    \item Se han interceptado comunicaciones entre la aplicación móvil y el robot mediante Wireshark, revelando tráfico en texto plano que permite replicar comandos sin autenticación.
  \end{itemize}

  \item \textbf{Observaciones y recomendaciones:} El sistema, aunque con una estructura física robusta, presenta una arquitectura de red y control muy expuesta por defecto.
  \begin{itemize}
    \item Se recomienda desplegar perfiles de seguridad DDS mediante SROS2 para ROS 2, cifrar las comunicaciones HTTP con TLS 1.3 y establecer ACLs sobre los nodos accesibles.
    \item También se sugiere reforzar la autenticación en los puntos de entrada, eliminar credenciales embebidas y capacitar al personal en prácticas de seguridad.
  \end{itemize}
\end{itemize}

A raiz del trabajo realizado se han reportado y publicado los siguientes CVEs: CVE-2025-41108, CVE-2025-41109 y CVE-2025-41110 \cite{cve-vision60}. Además se han creado dos vídeos demostración que permiten visualizar la problemática reportada en los CVEs mencionados \cite{video-collision, video-dos}.  

\subsection{Caso 2: Unitree A1}

\begin{figure}[htbp]
    \centering
    \includegraphics[width=0.5\linewidth]{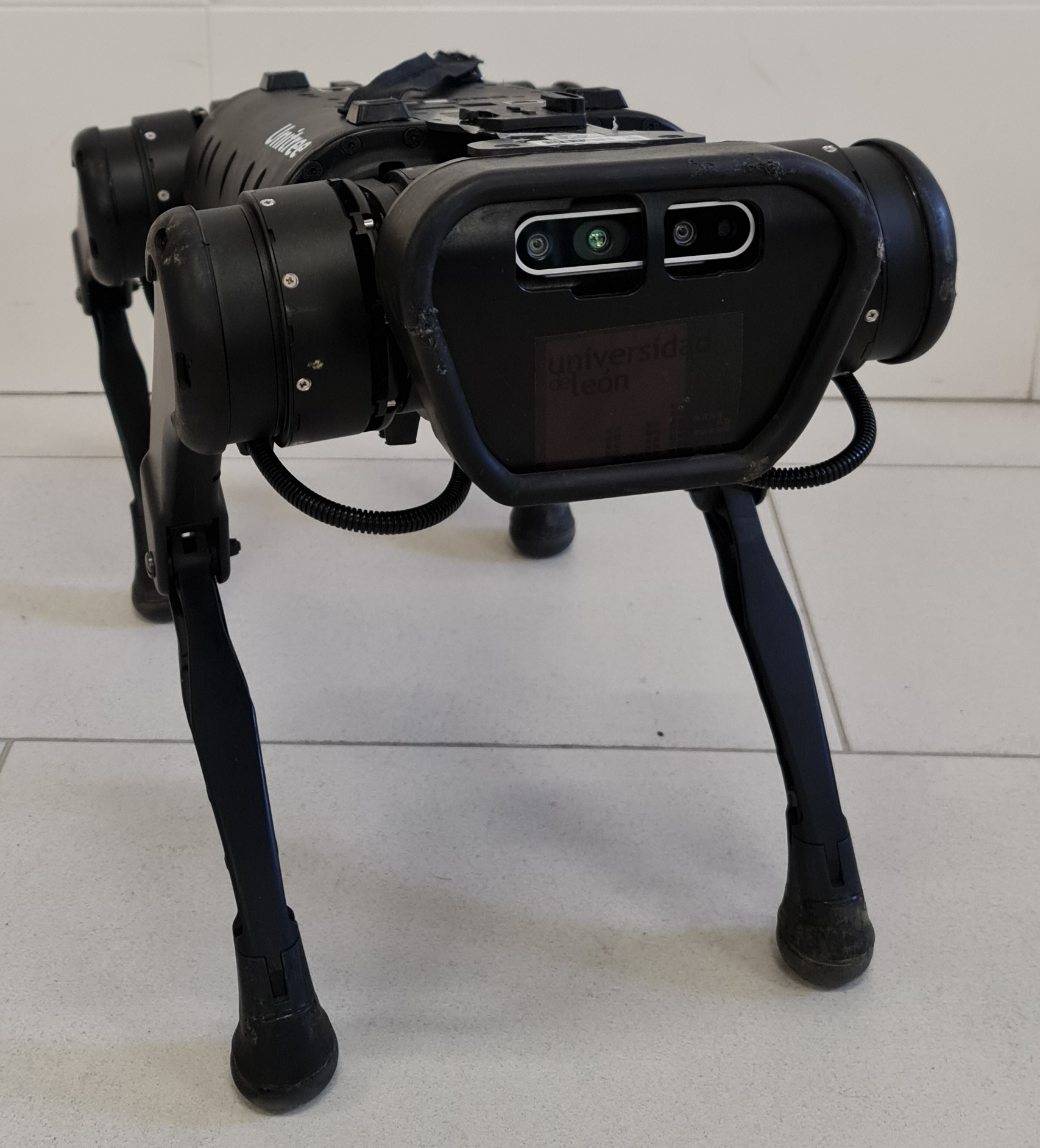}
    \caption{Unitree A1}
    \label{fig:tuercas}
\end{figure}

El Unitree A1 (ver Figura~\ref{fig:tuercas}) es un robot cuadrúpedo ligero y ágil, ampliamente utilizado en investigación y educación. Su arquitectura (ver Figura~\ref{fig:arquitectura_unitree}) es propietaria y no hace uso de un middleware conocido de control como el Vision 60 que utiliza ROS 2. El contenido de esta sección se corresponde con el análisis realizado en \cite{paperUnitreeA1}.

\begin{itemize}
  \item \textbf{Recopilación de información:} El primer paso del análisis ha consistido en una evaluación del entorno de ejecución, las aplicaciones de usuario y las conexiones de red empleadas (ver Figura~\ref{fig:conexiones_unitree}).
  \begin{itemize}
    \item El sistema operativo identificado es Ubuntu 16.04 con ROS  Kinetic, sin parches recientes.
    \item Se han inspeccionado los servicios activos: HTTP para el panel de administración, Wi-Fi en modo AP abierto, y un protocolo binario propietario transmitido sobre UDP.
    \item En último lugar, se ha procedido al análisis estático y dinámico de la aplicación oficial para Android (APK), empleando herramientas como JADX, APKTool y Frida.
  \end{itemize}
  \begin{figure}[htbp]
    \centering
    \includegraphics[width=\linewidth]{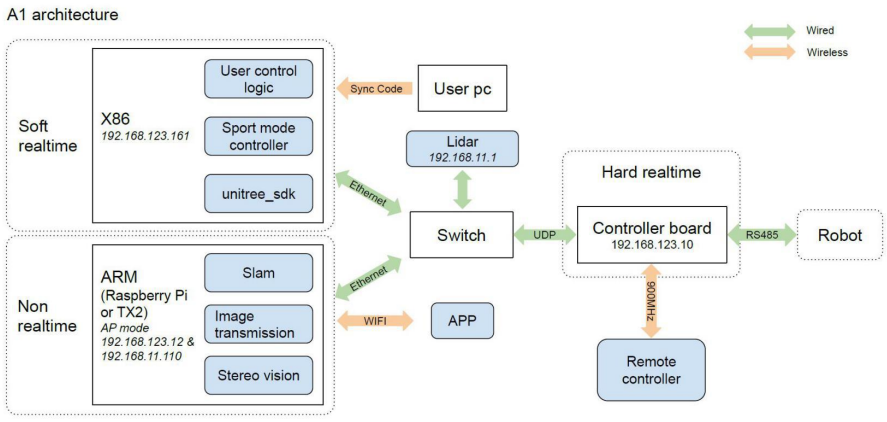}
    \caption{Arquitectura del Unitree A1}
    \label{fig:arquitectura_unitree}
\end{figure}

  \begin{figure}[htbp]
    \centering
    \includegraphics[width=0.5\linewidth]{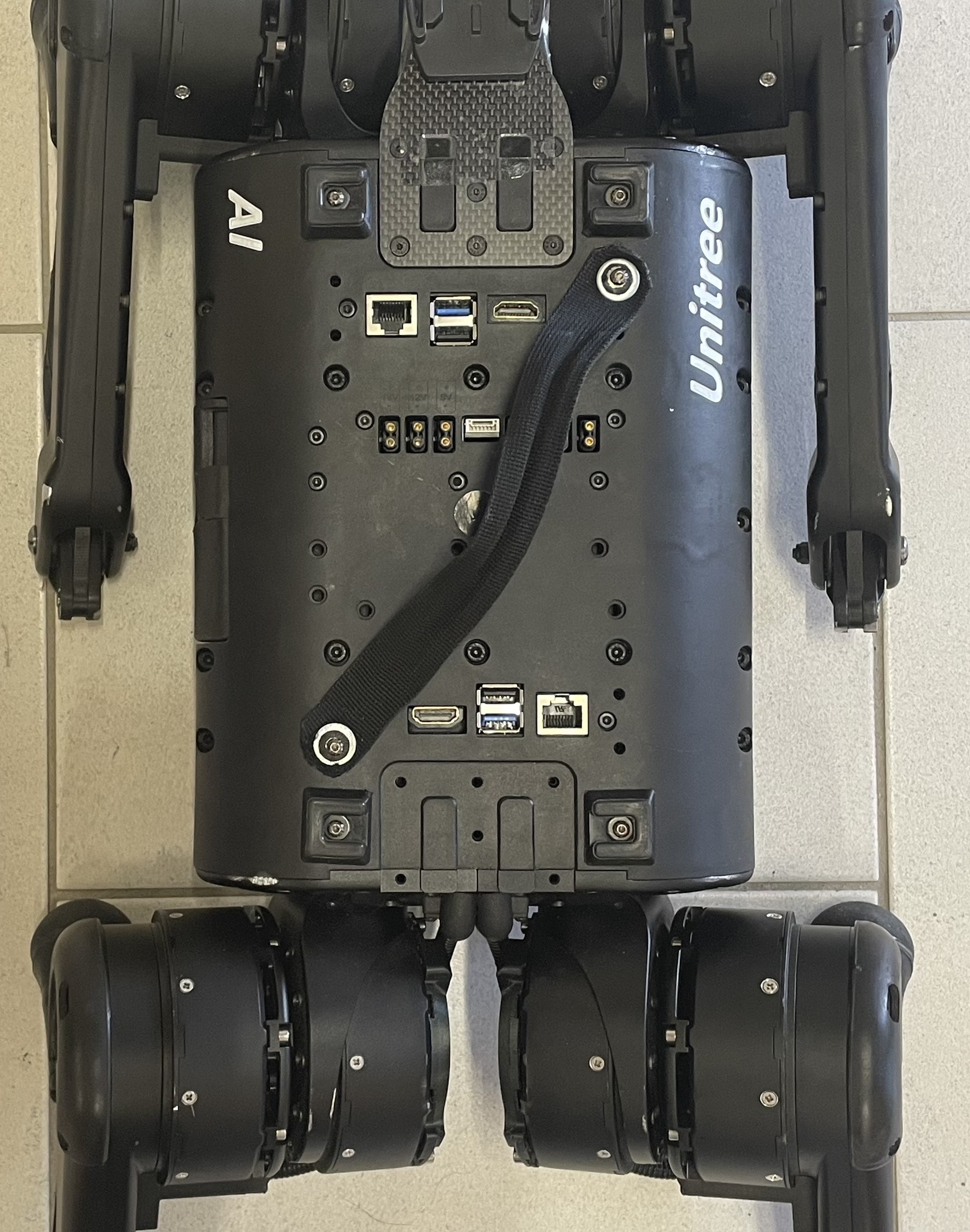}
    \caption{Conexiones Superiores del Unitree A1}
    \label{fig:conexiones_unitree}
\end{figure}

  \item \textbf{Análisis de vulnerabilidades:} El modelo de comunicación propietario presenta importantes deficiencias de diseño en términos de seguridad.
  \begin{itemize}
    \item Se han detectado múltiples comandos transmitidos sin cifrado ni autenticación, permitiendo su captura y reproducción.
    \item En la APK se han hallado credenciales codificadas en texto plano, direcciones IP por defecto y trazas de depuración activas.
    \item El servidor web desplegado en el robot no realiza validación ni implementa ningún tipo de autenticación.

  \end{itemize}

  \item \textbf{Explotación controlada:} Con las vulnerabilidades detectadas se han diseñado varios escenarios de prueba en entorno de red aislado.
  \begin{itemize}
    \item Se han replicado comandos de control a través de replay, provocando movimientos sin intervención del usuario.
    \item Se han inyectado comandos falsos utilizando herramientas como Scapy y hping3, sin que el sistema los detectara.
    \item Se ha realizado un ataque al servidor web con las imágenes de las cámaras del dispositivo, suplantando las imágenes legítimas por falsificaciones en tiempo real.
  \end{itemize}

  \item \textbf{Observaciones y recomendaciones:} El Unitree A1 presenta un diseño funcional pero muy débil en cuanto a ciberseguridad.
  \begin{itemize}
    \item Se recomienda rediseñar el protocolo de comunicación para incluir autenticación mutua y cifrado simétrico basado en claves negociadas.
    \item También resulta crucial eliminar las vulnerabilidades propias de sistemas antiguos como Ubuntu 16 e implementar una cadena de validación del firmware.
  \end{itemize}
\end{itemize}

A raíz del trabajo realizado se han reportado y publicado los siguientes CVEs: CVE-2023-3103 \cite{cve-unitree2} y CVE-2023-3104 \cite{cve-unitree1}.

\subsection{Caso 3: Brazo robótico UR3 de Universal Robots}

\begin{figure}[htbp]
\centering
\includegraphics[width=0.5\linewidth]{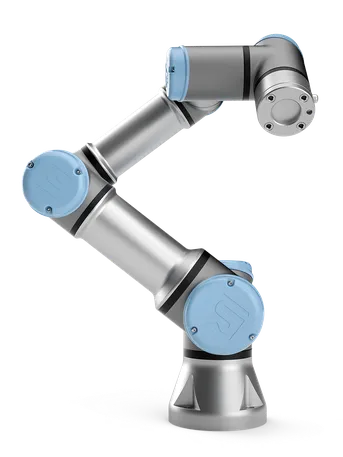}
\caption{UR3 de Universal Robots}
\label{fig:ur3}
\end{figure}

El UR3 (ver Figura~\ref{fig:ur3}) es un brazo robótico colaborativo desarrollado por Universal Robots, diseñado para tareas de ensamblaje de precisión, laboratorios y entornos industriales. Su arquitectura se basa en ROS Noetic sobre Ubuntu 18. Se permite su programación mediante una interfaz gráfica y una API accesible por red. Esta apertura facilita su integración pero también expone vectores de ataque si no se protegen adecuadamente los canales de comunicación.

\begin{itemize}
\item \textbf{Recopilación de información:} La fase inicial ha consistido en el mapeo del entorno de ejecución, los protocolos empleados y la arquitectura de comunicaciones entre el brazo, el controlador y el gripper.
\begin{itemize}
\item Se ha identificado el uso de un sistema Linux embebido en el controlador, con acceso remoto habilitado mediante SSH y una interfaz web en que utiliza el protocolo HTTP.
\item Es posible controlar el sistema vía scripts URScript a través de sockets TCP abiertos por defecto en el puerto 30002.
\item La comunicación entre el controlador y el gripper se realiza a través de un canal serie simulado o interfaz digital, sin mecanismos criptográficos de autenticación o cifrado.
\item Se analizaron paquetes transmitidos desde la GUI hacia el gripper, revelando comandos de bajo nivel fácilmente interceptables.
\end{itemize}

\item \textbf{Análisis de vulnerabilidades:} El modelo de control del UR3 expone múltiples superficies de ataque en sus interfaces abiertas y su diseño de control desacoplado.
\begin{itemize}
\item El protocolo de control del gripper que utiliza la primera versión de ROS, permite la inyección de comandos sin validación ni cifrado, lo que posibilita ataques del tipo man-in-the-middle.
\item Las configuraciones de fábrica permiten el acceso al API de scripting sin autenticación desde la red local, lo que aumenta el riesgo de accesos no autorizados.
\end{itemize}

\item \textbf{Explotación controlada:} Aprovechando las vulnerabilidades descubiertas, se desarrolló una prueba de concepto para interrumpir el funcionamiento del gripper.
\begin{itemize}
\item Mediante la herramienta nfqsed, se ha llevado a cabo la modificación de los paquetes de red intercambiados durante la comunicación entre topics de ROS, los cuales no están cifrados. Esta acción provoca un comportamiento no deseado en el funcionamiento del sistema robótico.

\item El ataque modifica los comandos enviados al gripper, impidiendo el cierre de sus pinzas incluso cuando este es solicitado por el programa original.
\item Esta técnica ha generado una denegación de servicio persistente sobre la funcionalidad de agarre, afectando directamente la utilidad del brazo en tareas de manipulación.
\end{itemize}

\item \textbf{Observaciones y recomendaciones:} Aunque el UR3 presenta una plataforma robusta en cuanto a mecánica y usabilidad, su arquitectura de comunicación presenta debilidades notables desde el punto de vista de la ciberseguridad.
\begin{itemize}
\item Se recomienda el cifrado de las comunicaciones internas entre el controlador y los periféricos, así como la implementación de autenticación mutua entre dispositivos.
\item El acceso al API por socket debe limitarse mediante autenticación fuerte y segmentación de red.
\item Además de los problemas encontrados, se ha realizado una búsqueda sobre posibles vulnerabilidades ya reportadas en este modelo de robot y se ha descubierto el CVE~\cite{cve-alias} el cual afirma que el controlador CB 3.1 del UR3  no cifra ni protege de ninguna manera los artefactos de propiedad intelectual instalados desde la plataforma UR+ de componentes de hardware y software (URCaps). Este fallo permite a los atacantes con acceso al robot o a la red de robots (en combinación con otros fallos) recuperar y exfiltrar fácilmente toda la propiedad intelectual instalada. Se ha comprobado que dicha vulnerabilidad continuaba activa en el momento del análisis.
\end{itemize}
\end{itemize}

\subsection{Caso 4: Pepper de Aldebaran Robotics}

\begin{figure}[htbp]
\centering
\includegraphics[width=0.2\linewidth]{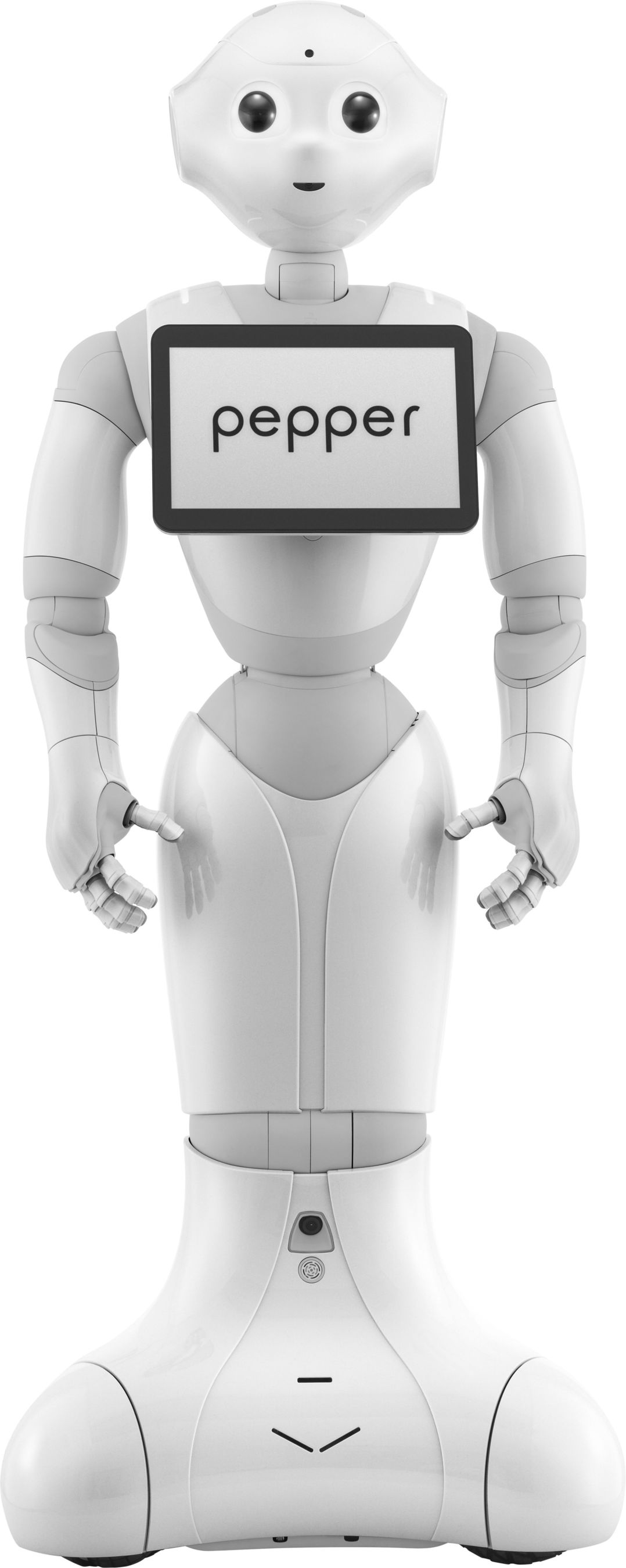}
\caption{Robot social Pepper de Aldebaran Robotics}
\label{fig:pepper}
\end{figure}

Pepper (ver Figura \ref{fig:pepper}) es un robot social semihumanoide fabricado por Aldebaran Robotics. Pepper está diseñado para interactuar con humanos en diferentes campos —educación, entretenimiento, hospitales y asistencia personal—, siendo capaz de reconocer caras y emociones básicas. El sistema permite interactuar con los usuarios a través de conversaciones y de una pantalla táctil. La arquitectura de Pepper está basada en un sistema operativo Linux embebido sobre el que se ejecuta el framework Naoqi 2.5.10.7, encargado de la gestión de los módulos sensoriales, motores y de comunicación.

El sistema cuenta con 20 grados de movimiento y está equipado con sensores táctiles, LED's, micrófonos, sensores infrarojos, bumpers, IMU, y cámaras 2D y 3D. Cuenta con capacidades de reconocimiento de voz en 15 idiomas distintos y proporciona una API para el desarrollo de aplicaciones personalizadas.

\begin{itemize}
    \item \textbf{Recopilación de información:} Durante esta primera fase se ha reunido información sobre la plataforma robótica, incluyendo el tipo de hardware y sensores que integra, el sistema operativo y los servicios que ejecuta.
    
    \begin{itemize}
        \item Se ha identificado el uso de un sistema operativo basado en Linux  con acceso remoto mediante SSH y una interfaz web que utiliza el protocolo HTTP sin cifrado en las comunicaciones 
        \item Se han identificado un total de 13 puertos abiertos con diferentes versiones de servicios antiguos.
        \item Se ha observado como en el puerto 9559  admiten conexiones sin autenticación
    \end{itemize}

    \item \textbf{Análisis de vulnerabilidades:} Tras las pruebas realizadas se han identificado las siguientes vulnerabilidades en la plataforma:
    \begin{itemize}
        \item Se ha detectado que la API implementada por Pepper y que permite un control absoluto del sistema no realiza ningún tipo de autenticación al tratar de acceder a la misma.
        \item El servidor web utilizado por el robot utiliza una comunicación HTTP sin encriptación.
        \item El sistema robótico no incluye mecanismos para prevenir ataques de fuerza bruta sobre sus servicios SSH.
    \end{itemize}

    \item \textbf{Explotación controlada:} A partir de las vulnerabilidades detectadas en la fase anterior se han establecido una serie de escenarios de prueba.
    \begin{itemize}
        \item Se ha accedido a la API del sistema a través del puerto 9559 a través de un código desarrollado en Python —los lenguajes Java y C++ también son soportados— y encontrándose en la misma red que el robot.
        \item Se ha realizado un sniffing del tráfico del servidor web y se han obtenido las credenciales del mismo.
        \item Se ha empleado la herramienta de código libre Hydra para verificar la vulnerabilidad del servidor SSH ante ataques de fuerza bruta.
    \end{itemize}

    \item \textbf{Observaciones y recomendaciones:} El robot Pepper presenta una arquitectura orientada a la interacción social, pero con una exposición significativa desde el punto de vista de la ciberseguridad. Se ha comprobado que el sistema mantiene múltiples servicios activos con versiones obsoletas, ausencia de cifrado en las comunicaciones y falta de autenticación en la API principal.
    \begin{itemize}
        \item Se recomienda restringir el acceso a la API (puerto 9559) mediante un sistema de  autenticación robusto.

        \item Es esencial migrar el servicio web a HTTPS mediante la implementación de TLS 1.3, asegurando el cifrado de las credenciales transmitidas.

        \item Debe deshabilitarse el acceso SSH por defecto o limitarse a cuentas específicas con autenticación por clave pública, así como establecer mecanismos de protección contra ataques de fuerza bruta (fail2ban, limitación de intentos, etc.).

        \item Se aconseja actualizar el sistema operativo y los servicios dependientes para mitigar vulnerabilidades conocidas en versiones antiguas.

        \item Se sugiere integrar políticas de monitorización y registro para la detección temprana de accesos no autorizados y reforzar las medidas de seguridad física sobre interfaces de red.

    \end{itemize}

\end{itemize}

\subsection{Comparativa de hallazgos}

Con el fin de facilitar el análisis transversal de los tres casos de estudio, la Tabla~\ref{tab:comparativa_casos_3} presenta una síntesis comparativa de los aspectos técnicos más relevantes observados en cada plataforma. Este resumen permite identificar patrones comunes de exposición, así como vulnerabilidades específicas asociadas al diseño, arquitectura de red y mecanismos de control. La comparativa también sirve como base para priorizar acciones de mitigación en sistemas robóticos ciberfísicos de naturaleza similar.

\newpage
\begin{landscape}
\begin{table}[!htbp]
\caption{Comparativa de resultados entre Vision 60, Unitree A1, UR3 y Pepper}
\label{tab:comparativa_casos_3}
\begin{center}
\begin{tabularx}{\linewidth}{|X|X|X|X|X|}
\hline
\textbf{Fase de auditoría} & \textbf{Vision 60 (Ghost Robotics)} & \textbf{Unitree A1 (Unitree Robotics)} & \textbf{UR3 (Universal Robots)} & \textbf{Pepper (Aldebaran Robotics)} \\
\hline
Sistema operativo y middleware & Ubuntu 18.04 + ROS 2 Dashing & Ubuntu 16.04 + ROS Kinetic & Ubuntu 18.04 + ROS Noetic & Linux + Naoqi 2.5.10.7 \\
\hline
Interfaces expuestas & SSH, HTTP sin TLS, ROS DDS, Wi-Fi abierto, Ethernet & HTTP sin autenticación, UDP binario, Wi-Fi sin cifrado & SSH, HTTP, API por sockets TCP (URScript), puerto 30002 expuesto & SSH, HTTP sin cifrar y API en puerto 9559 sin autenticación\\
\hline
Nivel de cifrado & Inexistente en consola web y ROS 2 DDS & Inexistente en comandos UDP y app móvil & Sin cifrado en canal controlador-gripper, tráfico no autenticado & Inexistente \\
\hline
Principales vectores de ataque & Consola sin autenticación, credenciales en app, DoS en ROS 2 & Replay y spoofing, comandos inseguros, análisis de APK & MITM al gripper, denegación de servicio por manipulación de comandos & Falta de autenticación en la API; servidor web sin cifrado; ausencia de mecanismos anti-fuerza bruta en SSH. \\
\hline
Impacto potencial & Modificación remota del comportamiento, pérdida de control autónomo & Pérdida de control, ejecución no autorizada de acciones físicas & Interrupción total de la función de agarre, pérdida de funcionalidad operativa & Compromiso total del sistema y robo de credenciales en texto plano.\\
\hline
Medidas recomendadas & Cifrado TLS, SROS2, segmentación de red, políticas de acceso físico & Rediseño de protocolo, cifrado simétrico, revisión del firmware y autenticación & Autenticación mutua, cifrado en canal gripper, control de acceso a APIs & Cifrar comunicaciones (TLS 1.3), reforzar autenticación, actualizar sistema operativo y monitorizar accesos  \\
\hline
\end{tabularx}
\end{center}
\end{table}
\end{landscape}

\newpage

\section{Taxonomía de amenazas en sistemas autónomos}

Los sistemas autónomos, como robots móviles, drones o vehículos sin conductor, constituyen una clase particular de sistemas ciberfísicos que integran de forma estrecha componentes computacionales, sensores, actuadores y canales de comunicación en tiempo real. Esta integración les permite operar de manera autónoma en entornos dinámicos y no estructurados, pero también los convierte en objetivos especialmente vulnerables a ataques que pueden originarse desde múltiples vectores.

A diferencia de los sistemas puramente digitales, la seguridad en sistemas autónomos no puede abordarse de forma aislada ni parcial. Un fallo en una capa aparentemente secundaria —como un canal de comunicación inseguro o un sensor mal configurado— puede desencadenar consecuencias físicas directas o comprometer decisiones críticas. Esta interdependencia entre el mundo físico y el digital obliga a considerar simultáneamente amenazas tanto lógicas como físicas, en escenarios locales y remotos.

Con el objetivo de facilitar el análisis, clasificación y priorización de riesgos, se propone una taxonomía de amenazas específica para sistemas autónomos, organizada en tres grandes dominios:

\begin{itemize}
    \item \textbf{Amenazas a nivel de hardware}, asociadas al acceso físico, la manipulación de componentes y la presencia de troyanos embebidos.
    \item \textbf{Amenazas a nivel de software y firmware}, centradas en la explotación de servicios vulnerables, configuraciones inseguras y código malicioso.
    \item \textbf{Amenazas en las comunicaciones}, donde se compromete la integridad, confidencialidad o disponibilidad de los canales de transmisión de datos.
\end{itemize}

En los siguientes apartados se describen con mayor detalle las amenazas asociadas a cada uno de los dominios.

\subsection{Amenazas a nivel de hardware}

Los componentes físicos de los sistemas autónomos representan una superficie de ataque crítica, especialmente en entornos donde los dispositivos operan sin supervisión directa o sin medidas de protección física adecuadas. A diferencia de los ataques puramente digitales, las amenazas a nivel de hardware permiten comprometer el sistema desde sus cimientos, afectando sensores, buses de datos, controladores o incluso el propio procesador. Entre las amenazas más relevantes en este ámbito se encuentran:

\begin{itemize}
  \item \textbf{Manipulación física}: el acceso directo al dispositivo puede permitir la modificación de sensores, actuadores o módulos de comunicación. Estos ataques pueden incluir el reemplazo de componentes por equivalentes maliciosos, la inserción de dispositivos intermediarios (como keyloggers físicos o sniffer USB), la reconfiguración de sensores mediante campos electromagnéticos o, incluso, el daño físico a los componentes electrónicos del sistema a través de dispositivos como el USB Killer \cite{angelopoulou2019killing}. Este tipo de amenaza es especialmente relevante en robots desplegados en exteriores, instalaciones industriales o entornos militares, donde la protección perimetral es limitada.

  \item \textbf{Exposición de interfaces de entrada/salida}: muchos sistemas embebidos mantienen activas interfaces de depuración como UART, JTAG, SPI o I2C, así como puertos estándar como USB o Ethernet. Si estas interfaces no están deshabilitadas, protegidas físicamente o autenticadas, pueden permitir acceso directo a la memoria del sistema, modificar configuraciones, o incluso cargar código malicioso sin necesidad de romper esquemas de cifrado. El uso de conectores estándar y placas comerciales facilita este tipo de ataques en entornos donde el acceso físico no está restringido.

  \item \textbf{Troyanos de hardware}: estos ataques implican la inserción intencionada de componentes maliciosos en el circuito del sistema, ya sea en fase de diseño, fabricación o mantenimiento. Pueden consistir en chips modificados, microcontroladores clonados con funciones ocultas, o puertas traseras físicas embebidas. Esta amenaza está especialmente presente en cadenas de suministro internacionales, donde la verificación del origen y funcionalidad real de los componentes es compleja. La presencia de troyanos puede pasar desapercibida durante años hasta ser activada remotamente o por una condición específica del entorno.
\end{itemize}

\subsection{Amenazas a nivel de software y firmware}

El software y el firmware constituyen el núcleo lógico de los sistemas autónomos. Comprenden desde el sistema operativo y los controladores hasta los entornos de middleware (como ROS/ROS 2), pasando por servicios de red, bibliotecas de terceros y aplicaciones de control. Estas capas son altamente vulnerables a ataques que explotan errores de programación, configuraciones inseguras o dependencias desactualizadas. Las amenazas más relevantes incluyen:

\begin{itemize}
  \item \textbf{Denegación de servicio (DoS/DDoS)}: los sistemas autónomos suelen contar con recursos limitados, por lo que una sobrecarga de CPU, memoria o red puede provocar bloqueos completos o inestabilidad. Estos ataques pueden dirigirse a servicios internos (por ejemplo, demonios ROS o servidores web embebidos), o utilizar tráfico externo malicioso para saturar los canales de comunicación o agotar los recursos computacionales.

  \item \textbf{Ejecución arbitraria de código y explotación remota}: vulnerabilidades en servicios en ejecución (por ejemplo, puertos abiertos, APIs mal protegidas...) pueden permitir a un atacante remoto ejecutar código malicioso. Este código puede alterar el comportamiento del sistema, comprometer funciones críticas o establecer persistencia a través de puertas traseras.

  \item \textbf{Rootkits y malware}: estos programas están diseñados para ocultarse y mantener el control sobre el sistema, incluso después de reinicios o actualizaciones. En un robot autónomo, la presencia de un rootkit puede facilitar el espionaje continuo, el sabotaje de tareas o la recopilación de datos sensibles sin ser detectado por los sistemas de supervisión.

  \item \textbf{Desbordamientos de búfer (buffer overflow)}: esta técnica clásica de explotación permite sobrescribir partes de la memoria para redirigir la ejecución del programa o escalar privilegios. En sistemas embebidos, donde el control de memoria suele ser manual o limitado, sigue siendo una amenaza muy vigente.

  \item \textbf{Ransomware y spyware}: aunque inicialmente dirigidos a sistemas tradicionales, estas amenazas han migrado hacia dispositivos embebidos e IoT. El ransomware puede cifrar configuraciones críticas o datos de operación del robot \cite{ransomRobotics}, mientras que el spyware puede registrar comandos, localizaciones o capturas de sensores.

  \item \textbf{Ataques de cracking de contraseñas}: mediante diccionarios, fuerza bruta o ingeniería social, los atacantes pueden obtener credenciales de acceso a interfaces administrativas, APIs o servicios remotos. El uso de contraseñas por defecto o esquemas de autenticación débiles es un factor recurrente en este tipo de ataques.

  \item \textbf{Ingeniería inversa}: el análisis del firmware o de aplicaciones móviles asociadas al sistema puede revelar claves embebidas, tokens de acceso, algoritmos de control o configuraciones inseguras. Este tipo de análisis es especialmente preocupante cuando el software se distribuye sin mecanismos de ofuscación o verificación de integridad.
\end{itemize}

\subsection{Amenazas a nivel de comunicaciones}

Los sistemas autónomos dependen de canales de comunicación para interactuar con su entorno, recibir instrucciones, transmitir datos sensoriales y coordinarse con otras entidades. Estos canales pueden incluir redes Wi-Fi, enlaces 5G, protocolos propietarios o buses industriales. Dado su carácter distribuido y muchas veces inalámbrico, la comunicación representa una superficie de ataque crítica, especialmente vulnerable a interceptación, manipulación o denegación.

Las amenazas más relevantes en esta categoría incluyen:

\begin{itemize}
  \item \textbf{Ataques de repetición (Replay)}: consisten en la captura de mensajes válidos transmitidos previamente, que son reenviados en momentos posteriores para ejecutar acciones no autorizadas. Son especialmente peligrosos en protocolos sin mecanismos de verificación temporal de paquetes  o sistemas de verificación basados en CRC, como algunos despliegues simples de MQTT, ROS o controladores industriales \cite{fernandes2017iot}.

  \item \textbf{Ataques Man-in-the-Middle (MitM)} \cite{mitmROS}: el atacante se posiciona entre el sistema autónomo y su controlador o servidor, interceptando y alterando el contenido de la comunicación. Esto le permite modificar comandos, capturar credenciales, o suplantar mensajes legítimos sin ser detectado. Es una amenaza crítica cuando no se utiliza cifrado de extremo a extremo ni autenticación mutua.

  \item \textbf{Suplantación de identidad de dispositivos (Masquerading)}: se produce cuando un atacante simula ser un nodo legítimo del sistema, como un sensor, actuador o unidad de control. Esto puede permitir el envío de datos manipulados, la desactivación de subsistemas o la manipulación de decisiones autónomas. Es un riesgo común en redes mal segmentadas o con protocolos que carecen de autenticación robusta.

  \item \textbf{Denegación de servicio distribuida (DDoS)} \cite{dosRobotics}: este ataque se basa en saturar los recursos del sistema (ancho de banda, capacidad de procesamiento) mediante un gran volumen de peticiones distribuidas desde múltiples fuentes. En sistemas autónomos conectados a la nube o accesibles remotamente, esto puede bloquear completamente la operación del robot o impedir su comunicación con los servidores de control.

  \item \textbf{Ataques de ralentización (Slowloris, HTTP flood)}: variantes de DoS que explotan debilidades en la gestión de sesiones y conexiones, manteniendo abiertas peticiones incompletas durante largos periodos para agotar los recursos del servidor. Son especialmente efectivos contra interfaces web embebidas o servicios de control HTTP expuestos.

  \item \textbf{Interferencia inalámbrica (Jamming)}: se refiere a la emisión deliberada de señales electromagnéticas con el objetivo de interrumpir las comunicaciones inalámbricas del sistema. Puede provocar pérdida de conectividad, comportamientos erráticos o activación de protocolos de emergencia. Es una amenaza relevante en entornos militares, industriales o con presencia de actores hostiles \cite{yang2013wireless}.
\end{itemize}

\begin{figure}
    \centering
    \includegraphics[width=1\linewidth]{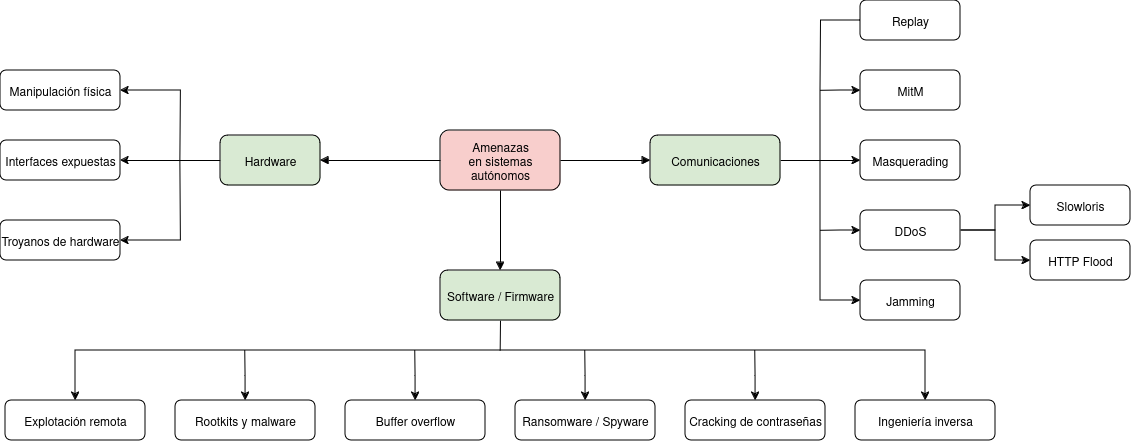}
    \caption{Taxonomía de amenazas en sistemas autónomos organizada por dominio}
    \label{fig:taxonomia_amenazas}
\end{figure}

Cada una de las amenazas descritas presenta riesgos específicos y mecanismos de ataque característicos, que deben analizarse en función del contexto operativo, el entorno de despliegue y la arquitectura del sistema. La Figura~\ref{fig:taxonomia_amenazas} muestra una representación visual de la taxonomía planteada.

\subsection*{Aplicación contextual de la taxonomía}

La taxonomía presentada permite clasificar y priorizar las amenazas en función del entorno operativo y las características específicas del sistema autónomo. Esta organización no solo facilita el diseño de auditorías dirigidas, sino que también orienta la aplicación de medidas de protección por dominio tecnológico, alineándose con el principio de defensa en profundidad.

Por ejemplo, en entornos industriales con red interna segmentada y dispositivos físicamente protegidos, las amenazas más relevantes tienden a concentrarse en el firmware y el hardware. En cambio, en sistemas móviles desplegados en espacios abiertos o expuestos a redes públicas —como drones, robots de reparto o plataformas de vigilancia— las amenazas a nivel de comunicaciones adquieren un peso mucho mayor.

La Tabla~\ref{tab:priorizacion_entornos_simple} resume de forma comparativa la criticidad, según la escala propuesta en \cite{incibe2023cvss}, de los distintos dominios de amenaza en diversos escenarios de aplicación.

\begin{table}[h]
\centering
\caption{Priorización relativa de dominios de amenaza por tipo de entorno (escala baja-media-alta)}
\label{tab:priorizacion_entornos_simple}
\begin{tabular}{|l|c|c|c|}
\hline
\textbf{Entorno de despliegue} & \textbf{Hardware} & \textbf{Software/Firmware} & \textbf{Comunicaciones} \\
\hline
Robótica industrial (en red cerrada) & Media & Crítica & Media \\ 
Robots móviles en entornos públicos & Media & Alta & Crítica \\ 
Drones de vigilancia o entrega & Media & Media & Crítica \\ 
Plataformas militares o críticas & Crítica & Crítica & Crítica \\ 
Entornos académicos o prototipos & Baja & Alta & Alta \\
\hline
\end{tabular}
\end{table}

Esta visión comparativa ayuda a enfocar los esfuerzos de evaluación y mitigación allí donde el impacto potencial y la probabilidad de ataque son más altos. En la sección siguiente se identifican las amenazas más críticas que se han observado de forma recurrente en auditorías realizadas sobre plataformas reales.

\section{Top-5 de amenazas en sistemas autónomos}

A partir del análisis de auditorías realizadas sobre distintas plataformas robóticas, así como de una serie de revisiones sistemáticas de literatura \cite{yaacoub2022robotics,humayed2017cyberphysical}, se han identificado cinco amenazas que destacan por su frecuencia, su facilidad de explotación y su impacto transversal en múltiples capas del sistema y que se enumeran a continuación:
\begin{itemize}
    \item \textbf{Ataques de denegación de servicio (DoS/DDoS)}
    \item \textbf{Secuestro del sistema (Hijacking)}
    \item \textbf{Inyección de datos falsos (False Data Injection)}
    \item \textbf{Ataques Man-in-the-Middle (MitM)}
    \item \textbf{Ataques físicos sobre interfaces expuestas}
\end{itemize}

Estas amenazas afectan tanto a dispositivos embebidos como a infraestructuras de control y comunicación, y pueden servir como punto de partida para ataques complejos en cadena. En las subsecciones siguientes se analiza cada una de las amenazas mencionadas, describiendo su funcionamiento, los vectores típicos de ataque y los riesgos específicos que plantean en el contexto de los sistemas autónomos.

\subsection{Ataques de denegación de servicio (DoS/DDoS)}

Los ataques de denegación de servicio representan una de las amenazas más comunes y efectivas contra sistemas autónomos, debido a su capacidad para inutilizar por completo un dispositivo sin necesidad de comprometer su integridad lógica o física. Este tipo de ataque consiste en saturar los recursos del sistema —ya sea el ancho de banda, la capacidad de procesamiento o las conexiones simultáneas— hasta el punto de impedir su funcionamiento normal.

En plataformas autónomas, como drones o robots móviles, la interrupción de la disponibilidad del sistema puede tener consecuencias críticas. La pérdida de conectividad puede hacer que el dispositivo entre en un estado de espera indefinido, active protocolos de emergencia, o incluso cause accidentes físicos si se encuentra en movimiento. A modo de ejemplo, un dron de vigilancia que mantiene su enlace con el operador mediante una conexión Wi-Fi podría quedar inutilizado si un atacante lanza un ataque de tipo flood contra su interfaz de red, impidiendo la recepción de comandos remotos y dejando al dispositivo en una situación vulnerable.

Desde el punto de vista técnico, los ataques DoS pueden realizarse mediante diversas técnicas: desde el envío masivo de paquetes ICMP o SYN para saturar las interfaces de red, hasta métodos más sofisticados como Slowloris, que agotan la capacidad de mantener conexiones abiertas. En su variante distribuida (DDoS), múltiples dispositivos (a menudo comprometidos como parte de una botnet) coordinan sus ataques, aumentando considerablemente el volumen de tráfico malicioso y dificultando su mitigación.

En el caso de sistemas autónomos embebidos, esta amenaza se ve amplificada por la escasa tolerancia a fallos de muchos dispositivos, la falta de mecanismos de protección perimetral y la dificultad de implementar soluciones de mitigación complejas en entornos con recursos limitados. Si no se dispone de sistemas de recuperación automática, aislamiento de procesos o filtrado inteligente de tráfico, un ataque DoS puede detener completamente la operativa del sistema, abriendo la puerta a amenazas adicionales mientras se mantiene fuera de servicio.

\subsection{Secuestro del sistema (Hijacking)}

El secuestro del sistema, o \emph{hijacking}, constituye una de las amenazas más críticas en el contexto de los sistemas autónomos, ya que permite a un actor malicioso asumir el control parcial o total del dispositivo, alterando su comportamiento de forma arbitraria. Esta amenaza suele materializarse tras la explotación de una vulnerabilidad en el sistema operativo, la red de comunicaciones o la interfaz de control, lo que permite al atacante emitir comandos como si fuera una fuente legítima.

El impacto de este tipo de ataques es especialmente grave en sistemas en movimiento, como vehículos autónomos o drones, donde la manipulación remota puede traducirse en colisiones, desvío de trayectorias o detención repentina del sistema en zonas sensibles. Una demostración paradigmática de esta amenaza tuvo lugar en 2015, cuando investigadores de seguridad lograron tomar el control de un Jeep Cherokee en circulación mediante el sistema de infoentretenimiento conectado a Internet. Desde una ubicación remota, fueron capaces de modificar parámetros como la velocidad o la dirección, evidenciando las implicaciones reales del hijacking en vehículos conectados \cite{miller2015jeep}.

En el caso de robots autónomos, esta amenaza puede surgir por la presencia de credenciales por defecto, puertos abiertos sin restricción, protocolos de control que carecen de autenticación o aplicaciones móviles mal diseñadas. La reutilización de claves, la falta de cifrado de los canales de comunicación o el uso de APIs sin control de acceso son vectores habituales de compromiso.

Una vez comprometido, el sistema puede ser utilizado no solo para sabotaje directo, sino también como plataforma para lanzar ataques a otros dispositivos conectados, exfiltrar datos o interceptar información sensorial. Por tanto, la prevención del hijacking debe abordarse como una prioridad, aplicando principios de defensa en profundidad que incluyan autenticación robusta, control de acceso basado en roles, auditoría de eventos y validación continua de firmware y software.

%%Revisado hasta aqui
\subsection{Inyección de datos falsos (False Data Injection)}

Los sistemas autónomos dependen críticamente de sensores para percibir su entorno, estimar su estado interno y tomar decisiones en tiempo real. Esta dependencia convierte a los canales sensoriales en un punto particularmente vulnerable frente a ataques de inyección de datos falsos (\emph{False Data Injection}, FDI) \cite{mo2012falseinjection}, en los que un atacante introduce información manipulada con el objetivo de inducir errores en el comportamiento del sistema.

A diferencia de los ataques que comprometen el sistema directamente, los ataques FDI buscan alterar la percepción del entorno, forzando al sistema a comportarse de forma incorrecta sin modificar directamente su lógica de control. Por ejemplo, un robot móvil que utiliza visión artificial para la navegación puede ser desviado de su trayectoria mediante imágenes manipuladas, generadas digitalmente o mediante técnicas físicas como proyección de patrones o uso de láser. Esta desviación puede dar lugar a colisiones, fallos en la misión o daños a infraestructuras o personas.

Además de la manipulación directa de sensores físicos, existen técnicas de inyección que actúan a nivel de red. Un atacante puede interceptar y modificar paquetes de datos sensoriales o suplantar nodos legítimos de la red, enviando valores erróneos que afectan al sistema de control. En plataformas distribuidas o basadas en middleware robóticos como ROS, donde los topics utilizados por los sensores no siempre están autenticados ni cifrados, este tipo de ataque puede ejecutarse con relativa facilidad.

El impacto de los ataques FDI va más allá de errores puntuales: pueden generar pérdida de confianza en los mecanismos autónomos, errores acumulativos difíciles de detectar y situaciones de riesgo físico. Su detección temprana requiere mecanismos de validación cruzada entre sensores  y detección de anomalías.

\subsection{Ataques Man-in-the-Middle (MitM)}

Los ataques del tipo Man-in-the-Middle (MitM) constituyen una amenaza significativa en los sistemas autónomos conectados, particularmente aquellos que dependen de comunicaciones remotas para recibir comandos o enviar datos sensoriales. En este tipo de ataque, el adversario se sitúa entre dos entidades que se comunican —por ejemplo, un robot y su estación de control— e intercepta, altera o suplanta los datos intercambiados sin que ninguna de las partes detecte la intrusión.

Este tipo de compromiso es especialmente peligroso en arquitecturas donde los datos transmitidos influyen directamente en las decisiones del sistema. Por ejemplo, el caso en el que un operador remoto envía comandos de navegación a un robot móvil, y un atacante logra interceptar estas órdenes, modificarlas, y reenviarlas con contenido falso. De este modo, el robot ejecuta instrucciones maliciosas sin que el operador tenga indicios de la manipulación.

El impacto de un ataque MitM es múltiple: permite al atacante acceder a información sensible (como credenciales, rutas o lecturas de sensores), modificar el comportamiento del sistema en tiempo real o incluso simular ser una entidad legítima dentro de la red. Este tipo de ataque puede ser especialmente silencioso y difícil de detectar si se implementa mediante herramientas especializadas o proxys transparentes.

Entre los factores que facilitan su éxito destacan el uso de protocolos de comunicación sin cifrado (como HTTP o Telnet), la ausencia de mecanismos de autenticación mutua entre los extremos de la comunicación y la utilización de redes inalámbricas abiertas o mal segmentadas. En sistemas autónomos desplegados en exteriores o integrados en infraestructuras críticas, estos ataques pueden tener consecuencias severas tanto desde el punto de vista operativo como de seguridad.

La prevención de ataques MitM requiere adoptar protocolos seguros (TLS, SSH, VPN), emplear certificados digitales verificados, y establecer políticas de autenticación robusta con validación de integridad y origen de los mensajes.

\subsection{Ataques físicos sobre interfaces expuestas}

La exposición física de interfaces externas en sistemas autónomos constituye una vulnerabilidad crítica que requiere especial atención en entornos donde la seguridad no ha sido priorizada frente a la funcionalidad o la interoperabilidad. Puertos USB, interfaces de red, tarjetas de expansión y paneles de control pueden ser aprovechados como vectores de ataque si carecen de mecanismos de autenticación o protección física adecuados. La ausencia de autenticación en estos puntos de acceso permite a un atacante ejecutar código malicioso, alterar configuraciones internas o exfiltrar información sensible sin necesidad de credenciales o privilegios especiales.

Entre los casos más relevantes se encuentran dispositivos maliciosos que explotan directamente estas interfaces. Por ejemplo, un USB Killer puede provocar daños físicos en la electrónica interna del sistema al inyectar una descarga eléctrica de alta tensión a través de un puerto USB, comprometiendo la integridad del hardware. De manera similar, un Rubber Ducky permite la ejecución automática de comandos y scripts maliciosos al ser detectado como un dispositivo de entrada legítimo, facilitando la instalación de malware, la modificación de parámetros críticos o la apertura de canales de comunicación no autorizados. Estos ejemplos ilustran cómo la falta de autenticación y control en interfaces físicas expuestas puede ser aprovechada para comprometer tanto la integridad como la disponibilidad de los sistemas autónomos.

La vulnerabilidad se amplifica cuando los puertos permanecen accesibles durante la operación normal del sistema, sin controles físicos o lógicos que restrinjan su uso. Esta situación no solo habilita la ejecución de código no autorizado, sino que también puede permitir la manipulación de sensores, la alteración de registros críticos y la interrupción de la operación de actuadores, aumentando significativamente el riesgo de fallos operativos o accidentes en entornos autónomos.

\section{Estrategias de defensa para amenazas en sistemas autónomos}

La protección de sistemas autónomos frente a ciberamenazas exige una estrategia integral que combine medidas preventivas, capacidades de detección y mecanismos eficaces de respuesta. A diferencia de los entornos puramente digitales, estas plataformas operan en el mundo físico, lo que amplifica las consecuencias de un ataque y requiere enfoques de seguridad multidimensionales.

Dada su exposición a redes inseguras, sensores manipulables y procesos autónomos de decisión, los sistemas ciberfísicos deben incorporar defensas que respondan a su arquitectura particular, sus limitaciones embebidas y su entorno de operación. Las soluciones propuestas deben ser compatibles con sus restricciones en términos de cómputo, energía, comunicaciones y movilidad.

En esta sección se presentan un conjunto de estrategias de mitigación estructuradas en torno a las cinco amenazas más relevantes identificadas en auditorías previas. Estas medidas combinan herramientas tecnológicas específicas, prácticas recomendadas de configuración y principios de gestión de seguridad, con el objetivo de reducir la superficie de ataque, aumentar la resiliencia operativa y dificultar la explotación de vulnerabilidades.

\subsection{Defensa contra ataques de denegación de servicio (DoS/DDoS)}

La mitigación de ataques de denegación de servicio en sistemas autónomos requiere una combinación de mecanismos de defensa perimetral, supervisión interna y segmentación funcional. El objetivo es garantizar la disponibilidad operativa del sistema incluso frente a intentos de saturación de recursos computacionales o de red.

Una de las medidas fundamentales consiste en el uso de firewalls embebidos o gateways inteligentes que filtren el tráfico entrante, bloqueando conexiones no autorizadas, paquetes malformados o patrones de tráfico sospechosos. En arquitecturas basadas en middleware robótico, como ROS/ROS 2, es recomendable aislar las interfaces críticas mediante políticas de red específicas que limiten el acceso a topics sensibles y reduzcan la exposición de nodos clave.

Complementariamente, la integración de sistemas de prevención de intrusiones (IPS), como Snort o Suricata, en puntos de entrada de la red —por ejemplo, routers industriales o dispositivos edge— permite detectar en tiempo real firmas de ataques DoS conocidos, incluyendo inundaciones SYN, floods UDP o conexiones HTTP anómalas. Aunque estas herramientas suelen emplearse en sistemas tradicionales, versiones optimizadas pueden adaptarse a entornos embebidos con recursos limitados.

Otro eje de defensa relevante es la monitorización continua del uso de recursos internos del sistema. Soluciones ligeras que recopilen métricas de uso de CPU, consumo de memoria, tráfico de red o actividad de procesos permiten identificar comportamientos anómalos y activar respuestas automáticas. Estas respuestas pueden incluir la limitación de procesos, reinicio de servicios afectados o la reconfiguración temporal de interfaces comprometidas.

La aplicación combinada de estas estrategias contribuye no solo a prevenir ataques DoS, sino también a contener su impacto y facilitar una recuperación rápida, evitando la degradación funcional prolongada del sistema autónomo.

\subsection{Defensa contra ataques de secuestro del sistema (Hijacking)}

Evitar el secuestro del sistema en plataformas autónomas implica establecer mecanismos de control de acceso sólidos que garanticen que solo entidades legítimas pueden interactuar con los componentes críticos del sistema. Este tipo de protección debe estar presente tanto en las interfaces físicas como en las capas de software y red.

Una de las medidas más eficaces es la implementación de esquemas de autenticación robusta, basados en certificados digitales, claves públicas asimétricas o protocolos estándar como OAuth 2. Estos mecanismos permiten verificar de forma inequívoca la identidad de los controladores remotos, evitando que un atacante pueda emitir comandos desde una fuente no autorizada o suplantar a un operador legítimo.

La segmentación de red también juega un papel clave en la prevención del hijacking. Es recomendable separar los canales de control de los de monitorización o servicio, limitando el acceso a interfaces administrativas mediante el uso de listas blancas, VPNs o redes privadas virtualizadas. Este aislamiento debe complementarse con políticas de control de acceso basadas en el principio de menor privilegio, de modo que cada componente del sistema tenga permisos estrictamente necesarios para su función.

Además, la detección temprana de intentos de secuestro puede reforzarse mediante sistemas de detección de intrusiones (IDS) con capacidad de análisis de comportamiento. Estos sistemas permiten modelar patrones normales de operación y generar alertas cuando se detectan desviaciones significativas, como comandos fuera de contexto, cambios súbitos en la frecuencia de comunicación o accesos desde direcciones inusuales.

El enfoque debe ser preventivo, pero también reactivo: un sistema bien diseñado debe ser capaz de revocar credenciales comprometidas, desconectar interfaces en riesgo o degradar funcionalmente sus capacidades para preservar su integridad bajo ataque.

\subsection{Defensa contra ataques de inyección de datos falsos (FDI)}

La protección frente a ataques de inyección de datos falsos debe centrarse en preservar la integridad, autenticidad y coherencia de la información sensorial, ya que cualquier alteración en los datos de entrada puede desencadenar decisiones erróneas en sistemas autónomos. Dado que estos sistemas se basan en la percepción del entorno para actuar, incluso pequeñas manipulaciones pueden tener consecuencias críticas.

Una de las estrategias más eficaces es la incorporación de mecanismos de verificación de integridad, como el uso de funciones hash criptográficas (por ejemplo, SHA-256) o firmas digitales, que permiten comprobar que los datos recibidos no han sido modificados desde su origen. Estos mecanismos pueden aplicarse tanto a la información almacenada como a los mensajes intercambiados entre sensores y módulos de control.

Además, el uso de redundancia sensorial aporta una capa adicional de defensa. Al disponer de múltiples sensores que miden una misma magnitud, es posible implementar esquemas de votación o fusión sensorial que detecten valores inconsistentes o fuera de rango. Este enfoque, ampliamente adoptado en sectores como la aeronáutica o los vehículos autónomos, permite descartar lecturas anómalas y aumentar la fiabilidad del sistema global.

Otra medida fundamental consiste en cifrar las comunicaciones internas entre sensores, actuadores y controladores. Aunque a menudo se asume que los buses internos están protegidos por estar físicamente contenidos en el dispositivo, ataques documentados han demostrado la viabilidad de interceptar o modificar dichos canales.

Estas estrategias, combinadas con modelos de consistencia lógica y detección de anomalías basadas en comportamiento, permiten reducir significativamente el riesgo asociado a la manipulación de datos sensoriales.

\subsection{Defensa contra ataques Man-in-the-Middle (MitM)}

La defensa frente a ataques Man-in-the-Middle (MitM) en sistemas autónomos exige una protección sólida de los canales de comunicación, tanto en términos de confidencialidad como de integridad. Dado que estos sistemas dependen de enlaces remotos para recibir comandos, enviar telemetría o interactuar con servicios en la nube, cualquier manipulación en tránsito puede comprometer gravemente su comportamiento.

Una medida fundamental para evitar este tipo de ataques es el cifrado de extremo a extremo mediante protocolos modernos como TLS 1.3, configurado con autenticación mutua basada en certificados digitales X.509. Esta configuración garantiza que tanto el emisor como el receptor de la comunicación puedan verificar su identidad de forma recíproca, y que el canal sea resistente a interceptaciones y modificaciones. Su aplicación debe extenderse no solo a las conexiones de red tradicionales, sino también a APIs RESTful o interfaces de control remotas.

Complementariamente, es necesario validar la identidad de cada entidad participante en cada sesión establecida, más allá de la verificación del canal. Esto implica el uso de tokens de sesión con expiración limitada y mecanismos antifalsificación que aseguren que cada petición proviene de una fuente legítima y autorizada. Esta validación contextual puede incluir datos como la ubicación, el patrón de uso o el historial de comunicaciones previas.

En redes locales, donde los ataques MitM pueden aprovechar técnicas como el ARP spoofing o el envenenamiento de DNS, se recomienda adoptar contramedidas específicas. Herramientas como ARP guard, filtros de MAC estáticos o el despliegue de DNSSEC contribuyen a reforzar la seguridad del enrutamiento y dificultan la suplantación de nodos en entornos internos.

Estas estrategias deben integrarse de forma transversal en la arquitectura del sistema, garantizando que las comunicaciones sean seguras por defecto y que cualquier anomalía en el canal pueda ser detectada y gestionada de forma oportuna.

\subsection{Defensa contra ataques físicos sobre interfaces expuestas}

La exposición de interfaces físicas sin mecanismos de autenticación o control constituye un vector de ataque crítico en sistemas autónomos, por lo que su mitigación requiere un enfoque integral que combine protección física, validación de dispositivos y monitoreo activo. Implementar estas medidas desde la fase de diseño es esencial para reducir la superficie de ataque y limitar el impacto de accesos no autorizados.

En primer lugar, es fundamental restringir el acceso físico a puertos y conexiones sensibles mediante el uso de cerraduras, carcasas protegidas, sellos de seguridad y compartimentos bloqueables. Estas medidas impiden que atacantes puedan conectar dispositivos maliciosos, como USB Killer o Rubber Ducky, que aprovechan la ausencia de autenticación para ejecutar código, dañar hardware o extraer información crítica. Además, la segmentación de interfaces y la limitación del acceso a aquellas estrictamente necesarias para operación y mantenimiento contribuyen a minimizar la exposición de los sistemas.

Junto con la protección física, la implementación de autenticación de dispositivos es una estrategia clave. Cada interfaz externa debe validar la identidad del dispositivo conectado mediante certificados digitales, llaves seguras o mecanismos de cifrado que impidan la ejecución de comandos no autorizados. Esta autenticación asegura que únicamente hardware legítimo pueda interactuar con el sistema, reduciendo significativamente el riesgo asociado a la inserción de dispositivos maliciosos.

El monitoreo y la auditoría de las interfaces expuestas complementan estas medidas. Sistemas de registro de eventos, detección de conexiones inesperadas y alertas automáticas permiten identificar intentos de acceso no autorizado en tiempo real, facilitando una respuesta inmediata antes de que se produzca un daño crítico. Asimismo, es recomendable implementar pruebas de penetración periódicas y análisis de riesgo físico para detectar posibles vulnerabilidades en los puntos de acceso.

Finalmente, estas medidas deben integrarse siguiendo principios de seguridad por diseño y por defecto, considerando que la protección de interfaces físicas no es un complemento opcional, sino un componente fundamental de la arquitectura del sistema. La combinación de control físico, autenticación robusta de dispositivos y monitoreo continuo constituye una estrategia integral para prevenir ataques físicos sobre interfaces expuestas y garantizar la integridad, disponibilidad y confiabilidad de sistemas autónomos críticos.

\section{Conclusiones}

La creciente integración de sistemas autónomos en entornos industriales, urbanos y críticos exige una revisión profunda de los enfoques tradicionales de auditoría de seguridad. A lo largo de este trabajo se ha propuesto un marco metodológico adaptado a las particularidades de estos sistemas ciberfísicos, combinando una estrategia de auditoría por capas, una taxonomía estructurada de amenazas y un conjunto de medidas de mitigación basadas en la experiencia práctica.

La evaluación realizada ha permitido identificar cinco amenazas especialmente relevantes —ataques de denegación de servicio (DoS/DDoS), secuestro del sistema (hijacking), inyección de datos falsos (FDI), ataques Man-in-the-Middle (MitM) y los ataques físicos sobre interfaces expuestas— por su frecuencia, impacto potencial y facilidad de explotación. Estas amenazas atraviesan distintos niveles del sistema, desde sensores y buses hasta software de control y canales de comunicación. Como síntesis de los vectores de riesgo más relevantes, la Tabla~\ref{tab:resumen_amenazas} presenta una visión comparativa de las cinco amenazas críticas identificadas en auditorías de sistemas autónomos. Se incluyen descripciones, ejemplos representativos y su impacto, con el fin de facilitar su análisis y priorización.

\begin{table}[h]
\centering
\caption{TOP-5 de amenazas críticas en sistemas autónomos}
\label{tab:resumen_amenazas}
\begin{tabular}{|p{3cm}|p{5.5cm}|p{5.5cm}|}
\hline
\textbf{Amenaza} & \textbf{Descripción} & \textbf{Impacto principal} \\
\hline
DoS/DDoS & Saturación de recursos del sistema para interrumpir su funcionamiento. & Interrupción total del servicio, pérdida de control. \\
\hline
Hijacking & Toma de control del sistema tras explotar una vulnerabilidad. & Comportamiento malicioso, sabotaje, colisiones. \\
\hline
FDI & Manipulación de datos sensoriales para alterar decisiones del sistema. & Decisiones erróneas, fallos de misión, accidentes. \\
\hline
Ataques MitM & Intercepción y modificación de comunicaciones entre entidades legítimas. & Suplantación, pérdida de integridad, alteración remota del comportamiento. \\
\hline
Ataques físicos sobre interfaces expuestas & Acceso no autorizado a puertos y conexiones externas debido a la ausencia de mecanismos de autenticación. & Compromiso del hardware y software, manipulación de sensores/actuadores, exfiltración de datos, interrupción de operaciones. \\
\hline
\end{tabular}
\end{table}

La protección frente a estas amenazas requiere una aproximación integral que combine medidas preventivas, capacidades de detección y mecanismos de recuperación. Factores como la conectividad, las restricciones embebidas y los requisitos de operación en tiempo real condicionan qué soluciones pueden aplicarse en cada entorno. En consecuencia, no existen medidas universales, sino estrategias que deben ser seleccionadas y adaptadas al contexto específico del sistema autónomo.

En sistemas con alta criticidad operativa, como robots industriales, vehículos conectados o plataformas móviles en espacios públicos, se recomienda alinear la arquitectura de seguridad con marcos normativos reconocidos, tales como IEC 62443 \cite{IEC62443}, ISO/SAE 21434 \cite{iso21434} o el marco NIST para IoT \cite{nist2021iotframework}. Esta alineación permite incorporar principios de defensa en profundidad, trazabilidad de riesgos y validación formal de contramedidas.

A partir del análisis de las amenazas anteriores, la Tabla~\ref{tab:resumen_mitigaciones} sintetiza las principales líneas de defensa recomendadas para cada caso. Estas estrategias combinan medidas tecnológicas y organizativas, adaptables a las limitaciones y requisitos específicos de cada sistema autónomo.

\begin{table}[h]
\centering
\caption{Resumen de estrategias de defensa por amenaza}
\label{tab:resumen_mitigaciones}
\begin{tabular}{|p{3.2cm}|p{9.8cm}|}
\hline
\textbf{Amenaza} & \textbf{Estrategias principales de defensa} \\
\hline
DoS/DDoS & Filtrado de tráfico con firewalls embebidos, segmentación de red en ROS/ROS 2, integración de sistemas IPS, monitorización de recursos y activación de respuestas automáticas. \\
\hline
Hijacking & Autenticación basada en certificados, control de acceso con privilegios mínimos, segmentación de canales administrativos, detección de anomalías mediante IDS con perfilado de comportamiento. \\
\hline
FDI & Verificación de integridad con hashes o firmas, redundancia sensorial con votación mayoritaria, cifrado de buses de comunicación, detección de inconsistencias mediante validación cruzada. \\
\hline
MitM & Cifrado de extremo a extremo con TLS 1.3 y autenticación mutua, validación de tokens de sesión, protección contra ARP spoofing y DNS poisoning con ARP guard y DNSSEC. \\
\hline
Ataques físicos sobre interfaces expuestas & Restricción de acceso físico mediante cerraduras y carcasas, autenticación de dispositivos conectados (certificados, llaves seguras), monitoreo y auditoría de interfaces externas. \\
\hline
\end{tabular}
\end{table}

Si bien la metodología presentada ofrece una base sólida para auditar y proteger sistemas autónomos frente a ciberamenazas, existen algunas limitaciones inherentes al alcance del trabajo que abren la puerta a líneas futuras de investigación:

\begin{itemize}
  \item \textbf{Validación empírica ampliada}: aunque se ha demostrado la aplicabilidad del procedimiento mediante los casos de prueba realizados, sería conveniente replicar la metodología en un conjunto más amplio y diverso de plataformas (vehículos autónomos, robots sociales, sistemas aéreos no tripulados, etc.) para evaluar su generalización.
  \item \textbf{Automatización de auditorías}: la integración de herramientas automáticas de escaneo y análisis adaptadas a middleware robótico como ROS 2 aún está en una fase incipiente. Futuras líneas podrían centrarse en el desarrollo de frameworks automatizados de auditoría de seguridad específicos para sistemas ciberfísicos.
  \item \textbf{Criterios de priorización dinámica}: la priorización de amenazas actualmente se basa en experiencia previa y criterios estáticos. Sería valioso investigar técnicas basadas en aprendizaje automático o análisis de riesgo adaptativo que permitan ajustar la auditoría en función del contexto operacional del sistema.
  \item \textbf{Inclusión de aspectos legales y normativos}: si bien se han citado estándares relevantes, una integración sistemática de requisitos regulatorios (p. ej., NIS2, RGPD, ISO/IEC 27001) en cada fase de la auditoría aportaría valor a entornos con exigencias de cumplimiento normativo.
  \item \textbf{Ciberresiliencia y recuperación}: el trabajo se ha centrado principalmente en la prevención y detección de amenazas. Futuros desarrollos deberían profundizar en estrategias de resiliencia post-ataque, incluyendo protocolos de recuperación automática, redundancia funcional y validación de integridad posterior a incidentes.
\end{itemize}

En conjunto, estas líneas apuntan a consolidar el procedimiento aquí descrito como una metodología robusta, extensible y práctica para la protección efectiva de plataformas autónomas en escenarios reales.

\section*{Agradecimientos}
Esta publicación es parte del proyecto TESCAC, financiado/a por la “Unión Europea NextGeneration-EU, Plan de Recuperación, Transformación y Resiliencia, a través de INCIBE”.

\newpage

\bibliographystyle{IEEEtran}
\bibliography{referencias}

\end{document}